\documentclass[journal]{IEEEtran}

\ifCLASSINFOpdf
  \usepackage[pdftex]{graphicx}
  \graphicspath{{img/}}
  \DeclareGraphicsExtensions{.pdf,.jpg,.png}
\else
\fi

\usepackage{amsmath}

\ifCLASSOPTIONcompsoc
  \usepackage[caption=false,font=normalsize,labelfont=sf,textfont=sf]{subfig}
\else
  \usepackage[caption=false,font=footnotesize]{subfig}
\fi

\usepackage{url}

\usepackage{graphbox}
\usepackage{booktabs}
\usepackage{siunitx}
\usepackage{arydshln}
\usepackage{multirow}
\usepackage[normalem]{ulem}
\usepackage[linesnumbered,ruled,vlined,longend,norelsize]{algorithm2e}
\usepackage{pifont}
\newcommand{\cmark}{\ding{51}}%
\newcommand{\xmark}{\ding{55}}%
\usepackage{todonotes}
\usepackage{textpos}
\usepackage{hyperref}

\makeatletter
\newcommand{\removelatexerror}{\let\@latex@error\@gobble}
\makeatother

\SetAlFnt{\scriptsize\rmfamily}

\begin{document}

\title{Procedural Generation of 3D Maps\\with Snappable Meshes}

\author{Rafael~C.~e~Silva,
        Nuno~Fachada,
        Diogo~de~Andrade, %
        and~Nélio~Códices
\thanks{All authors are with the School of Communication, Arts and Information
Technology at Lusófona University, Lisboa, Portugal e-mail:
nuno.fachada@ulusofona.pt}
\thanks{N. Fachada and D. de Andrade are with COPELABS, Lusófona University}
\thanks{N. Códices is with Instituto Superior Técnico, Lisboa, Portugal}
}

\maketitle

\begin{textblock*}{190mm}(-1cm,-6cm)
  \noindent \footnotesize The peer-reviewed version of this paper is
  published in IEEE Access at
  \href{https://doi.org/10.1109/ACCESS.2022.3168832}{\texttt{https://doi.org/10.1109/ACCESS.2022.3168832}}.
  This version is typeset by the authors and differs only in pagination and
  typographical detail.
\end{textblock*}

\begin{abstract}
  In this paper we present a technique for procedurally generating 3D maps using a
  set of premade meshes which snap together based on designer-specified visual
  constraints. The proposed approach avoids size and layout limitations, offering
  the designer control over the look and feel of the generated maps, as well as
  immediate feedback on a given map's navigability. A prototype implementation of
  the method, developed in the Unity game engine, is discussed, and a number of
  case studies are analyzed. These include a multiplayer game where the method was
  used, together with a number of illustrative examples which highlight various
  parameterizations and piece selection methods. The technique can be used as a
  designer-centric map composition method and/or as a prototyping system in 3D
  level design, opening the door for quality map and level creation in a fraction
  of the time of a fully human-based approach.
\end{abstract}

\begin{IEEEkeywords}
3D maps, computer games, designer-centric methods, layout, procedural content
generation (PCG)
\end{IEEEkeywords}

\IEEEpeerreviewmaketitle

\section{Introduction}

\IEEEPARstart{I}{n} this paper we describe a method for procedurally generating 3D maps
using visual constraints. The approach, termed \textit{snappable meshes}, consists of a system
of connectors with pins and colors---similar in concept to a jigsaw puzzle---which constrains
how any two map pieces (i.e., meshes) can snap together. Through the visual design and
specification of these connection constraints, and an easy-to-follow and fully explainable
\cite{zhu2018explainable} generation procedure, the method is accessible to game designers
and/or other non-experts in procedural content generation (PCG), artificial intelligence
(AI) or programming.

While the maps are procedurally generated, the technique grants the game designer considerable
influence over the final result. More specifically, the designer has complete control on the
modeling of individual meshes---including the placement and configuration of associated
connectors---and can achieve substantial customization on the generated maps through a small
number of intuitive parameters. At the same time, the technique avoids size and layout
limitations, common in grid-based approaches \cite{smelik2014survey}. Maps are generated almost
immediately and can be quickly validated for navigability. With a focus on fast iteration and
respect for existing development workflows \cite{lai2020towards}, the snappable meshes PCG
technique can be used in practical industry settings, while following academic
best practices such as transparency and reproducibility \cite{peng2011reproducible}. Further,
the technique can be easily tweaked and extended by programmers, since one of its core
components---the \textit{selection method}, discussed in detail in Section~\ref{sec:methods}---is
fully swappable, and implementing new ones is relatively simple. Four of these \textit{selection
methods} are described in this paper and included in the provided prototype implementation.

The snappable meshes PCG technique was initially developed for a multiplayer combat game,
created as a semester project at Lusófona University's Bachelor in Videogames---an
industry-focused, interdisciplinary game development degree \cite{fachada2020topdown}.
The decision of using PCG in this particular game project was made to promote its replayability,
requiring players to adapt to a new map on every match, keeping the experience from turning stale
once the combat loop is mastered. Given the technique's capability of creating general 3D maps
(i.e., not specific to action games), its core ideas were previously presented in a conference
\cite{esilva2020procedural}. The present paper extends that publication in several ways, making
the following contributions:

\begin{itemize}
    \item The technique is presented in a carefully formalized, fully reproducible and
        implementation-independent fashion. Furthermore, it is comprehensively classified and
        contextualized within the PCG state of the art, highlighting the advantages it
        brings with respect to explainability \cite{zhu2018explainable} and potential integration
        with industry workflows \cite{lai2020towards}.
    \item A new optional connection constraint is introduced in the form of a color compatibility
        layer, specified through a color matrix. Additionally, two navigability-focused map
        validation metrics are proposed and analyzed.
    \item A reference and standalone implementation of the technique, developed in the Unity
        game engine \cite{unity3d}, is presented and examined. This implementation is made available
        as free and open-source software, and the source code is thoroughly annotated. Further,
        it includes concrete solutions for issues considered separate from the technique itself,
        such as mesh overlap detection (i.e., collision avoidance) and specific issues related with
        validating map navigability.
    \item Using the prototype implementation, an in-depth analysis of the technique's generation
        and validation capabilities is undertaken, namely on how the various parameters and
        \textit{selection methods} influence the created maps, as well as the generation and
        validation times.
    \item A thorough discussion is presented on several aspects of the proposed technique, related
        to its designer-centric approach, applicability (i.e., in which scenarios could the method
        be more useful), limitations, as well as alternative uses and possible improvements.
\end{itemize}

The paper is organized as follows. In Section~\ref{sec:background}, we review related work
concerning the use of PCG for map design in computer games.
In Section~\ref{sec:methods}, we describe the snappable meshes PCG technique, namely the
components that make up the system, and classify the
method according to a well-known PCG taxonomy. A reference implementation of the method,
developed in the Unity game engine, is presented in Section~\ref{sec:impl}. Several case
studies are examined in Section~\ref{sec:casestudy}, including one where the method was used
for generating arenas for the aforementioned multiplayer combat game. A discussion of the
method, possible uses, limitations and potential improvements follows in
Section~\ref{sec:discussion}. Section~\ref{sec:conclusion} closes the paper, offering
some conclusions.

\section{Background}
\label{sec:background}

The use of PCG for creating maps and levels in computer games began in the 1970's with games
such as \textit{Beneath Apple Manor} and \textit{Rogue} \cite{aycock2016pcg}, and is unarguably the
most common use case of PCG in games \cite{yannakakis2018artificial}. Academic interest in this area
has been growing, with a number of important developments occurring in the last 15 years \cite{togelius2011search,togelius2016intro}.

PCG techniques are often fully autonomous, in the sense that the user simply performs some initial
configuration, such as defining parameters and/or output constraints, before content is generated.
Recently, mixed-initiative
content creation (MICC) techniques, in which a combination of human input and computer-assisted
PCG are used, have gained traction in game development in general and map design in particular
\cite{yannakakis2018artificial,liapis2016mixed}. However, the line that separates autonomous PCG
and MICC-PCG is ill-defined. For example, Yannakakis \& Togelius \cite{yannakakis2018artificial}
claim that ``the PCG process is autonomous when the initiative of the designer is limited to
algorithmic parameterizations before the generation starts''. Then, how to classify a PCG technique
that works according to this statement, but produces results which mainly depend on
designer-provided building blocks, such as the one presented in this paper? Although we do not
provide a definitive answer, this question guides the type of related works discussed in this
section. More specifically, we will focus on PCG methods and tools for creating maps and levels in
which the human designer has considerable influence on the look and feel of the generated output,
irrespective of whether the PCG algorithm tends to be more autonomous (i.e., the designer specifies
the parameters and possibly provides the building blocks) or works in a more MICC fashion, where
human and computer iteratively collaborate on the design process.

Tanagra is one such MICC tool for 2D platformers \cite{smith2010tanagra}, allowing human designers
to partially specify a level's geometry and pacing, leaving it up to the computer to fill in the
gaps. Using constraint programming, Tanagra guarantees that the generated levels are playable
when human-defined constraints are valid.

Occupancy-regulated extension (ORE) \cite{mawhorter2010ore} is a 2D geometry assembly algorithm
that, similar to the algorithm presented in this paper, (i) creates maps from premade pieces (or
chunks), and, (ii) delegates context and chunk selection to separate subroutines. However,
contrary to the snappable meshes technique, ORE utilizes the potential positions a player might
occupy in a chunk during play to determine which other chunks are adequate for combination, thus
requiring pieces of geometry annotated with gameplay information.

PCG can be used together with pre-established level design patterns to generate desirable
levels. Two proposals have employed a MICC approach, where the contribution from the level
designers is used in the optimization process of the generators' evolutionary
algorithm~\cite{baldwin2017mixed,walton2020mixed}. This way, their accumulated knowledge of
level design is fed into the generators, refining not only level layouts but also agent and
item placement on each iteration. This is accomplished by identifying and then reinforcing
the level design patterns in the heuristics, leading to both playable and enjoyable
experiences.

Sentient Sketchbook \cite{liapis2013sentient} is a MICC level design tool for strategy games,
where designers sketch maps while being presented with similar, more detailed alternatives.
These recommendations are presented in real time using evolutionary search and employing gameplay
metrics---such as balance, exploration, resources or navigability---as fitness dimensions.

A number of 3D map and level generation approaches have also been proposed. SuSketch
\cite{migkotzidis2021susketch}, aimed at generating first-person shooter levels, is one such case.
Like Sentient Sketchbook, the user provides the initial designs and the tool offers alternative
layouts and feedback on a number of gameplay metrics in a full MICC fashion. Additionally, SuSketch
also provides gameplay predictions which take into account different character classes.

FPSEvolver is a 3D MICC generation proposal in which the human contribution for level design
comes from the players themselves. It consists of a multiplayer shooter featuring a novel
grid-based interactive evolution approach for generating maps according to the players'
preferences \cite{olsted2015interactive}. Players vote on a selection of evolving scenarios,
with the goal of generating levels in accordance with what they consider to be a good map.

Oblige is a 3D level generator for the DOOM family of games \cite{apted2017oblige}.
It allows the level designer to set a number of parameters, such as level size and
approximate quantities of each type of monster, power-up and level section (e.g. outdoors,
caves, hallways, etc). Levels are created using shape grammars on a grid-based layout,
and are limited to a single floor---an inherent limitation of the DOOM family of games.

Butler et al.~\cite{butler2013mixed} explored MICC-based PCG in the context of macro level
design, where the focus is the entire player experience and not just single levels. Their
solution relies on a set of authoring tools where level designers can define progression
constraints that lead to the generation of a progression plan. The solution then generates
and/or validates the individual level designs against the progression plan.

Regarding recent commercial applications, map and/or level PCG has been employed for level
creation in a number of games, namely 2D roguelikes such as \textit{Spelunky Classic}
\cite{yu2009spelunky}, \textit{The Binding of Isaac: Rebirth} \cite{mcmillen2014thebinding}
and \textit{Enter the Gungeon} \cite{dodgeroll2016enter}. \textit{Spelunky} uses premade room
templates to fill out a grid. Rooms with specific characteristics such as top entrances and
bottom pits are considered when generating levels in order to create a valid path for the
player to traverse towards the end \cite{yu2016spelunky,kazemi2013spelunky}. In
\textit{The Binding of Isaac: Rebirth}, maps are created by connecting several rooms together,
fit into a grid \cite{florian2020isaacexp,cuttingroom2019isaac}. However, each
room may take more than one grid space, and each grid space it occupies can be connected to other
rooms on adjacent grid spaces. This allows for large rooms to connect to small rooms and
vice-versa. \textit{Enter the Gungeon} does not connect its rooms directly to one another,
employing a more complex algorithm to obtain the desired layout. It uses nodes and connectors,
placing different premade rooms as those nodes and afterwards creating corridors to join the
rooms for the final map layout \cite{boris2019gungeon}.

Generate Worlds \cite{dykeman2019introducing} is a commercial tool for creating 3D
maps from user-provided voxel tiles or blocks. It uses a very simple premise: two blocks
are only adjacently placed if they have the same color in all the places where they touch,
in a similar fashion to Wang tiles \cite{wang1961proving}, often used for image and
texture generation \cite{cohen2003wang,lagae2009wang}. If the user carefully designs
these blocks, the method is able to generate varied and interesting content, from dungeons
to landscapes. Generate Worlds builds on a previously proposed open-source tool
\cite{dykeman2017procedural}, and essentially solves the constraint satisfaction problem
of correctly tiling blocks according to their colors. The Wave Function Collapse (WFC)
family of algorithms, initially proposed by Maxim Gumin \cite{gumin2016wfc}, takes these
ideas further, requiring only one input example (e.g., an image) and a rule to decompose it
into blocks (e.g., tiles). For the purpose of image, map and level generation, WFC works
essentially on a grid, although it has been shown to handle graph representations for more
specific generation tasks, such as node placement in navigation meshes \cite{kim2019automatic}.
Nonetheless, gridless map generation has yet to be demonstrated. WFC has been gaining
momentum among game developers, and was recently analyzed and formalized by Karth and Smith
\cite{karth2021wavefunctioncollapse}, which also look into several of the technique's uses
and extensions since its inception. A recent tool, Tessera \cite{newgas2021tessera},
addresses several common issues with the original WFC implementation, allowing users to
experiment with the technique and several of its extensions within the Unity game engine.

The map generation technique proposed in this paper uses some of the ideas present in the
works discussed thus far, but follows its own distinct approach. For example, contrary to
the work of Butler et al. \cite{butler2013mixed}, the proposed method does not deal with
gameplay or inter-level balance, focusing instead on the layout of individual maps.
Furthermore, unlike evolutionary-based designer feedback approaches
\cite{baldwin2017mixed,walton2020mixed,liapis2013sentient,migkotzidis2021susketch},
the presented PCG technique relies on the designer to carefully model individual map pieces,
since one of the goals is to generate multiple quality maps, rather than refining one design.
Like Sentient Sketchbook \cite{liapis2013sentient} and SuSketch \cite{migkotzidis2021susketch},
the snappable meshes technique uses path finding for map evaluation; contrary to these tools,
however, the technique is less bound to specific game genres (at the cost of less detailed maps)
and offers a fully explainable generation process. Snappable meshes follows a constraint-based
constructive approach, similar to Generate Worlds \cite{dykeman2019introducing} and WFC
\cite{gumin2016wfc}, but is not bound to a grid, allowing free-form map generation. Indeed,
it is considerably more flexible than Generate Worlds, as it does not require pixel-perfect
colored block interfaces, and is simpler than WFC, which several users found difficult to
grasp and refactor \cite{karth2021wavefunctioncollapse}. Further, unlike many of the works
discussed here, the proposed technique is not a specific tool designed for certain use cases,
but a generic and easy-to-understand procedure which can be integrated in various designer
tools and workflows.

\section{The Snappable Meshes PCG Technique}
\label{sec:methods}

\begin{figure}[!t]
  \centering
  \includegraphics[width=1\linewidth]{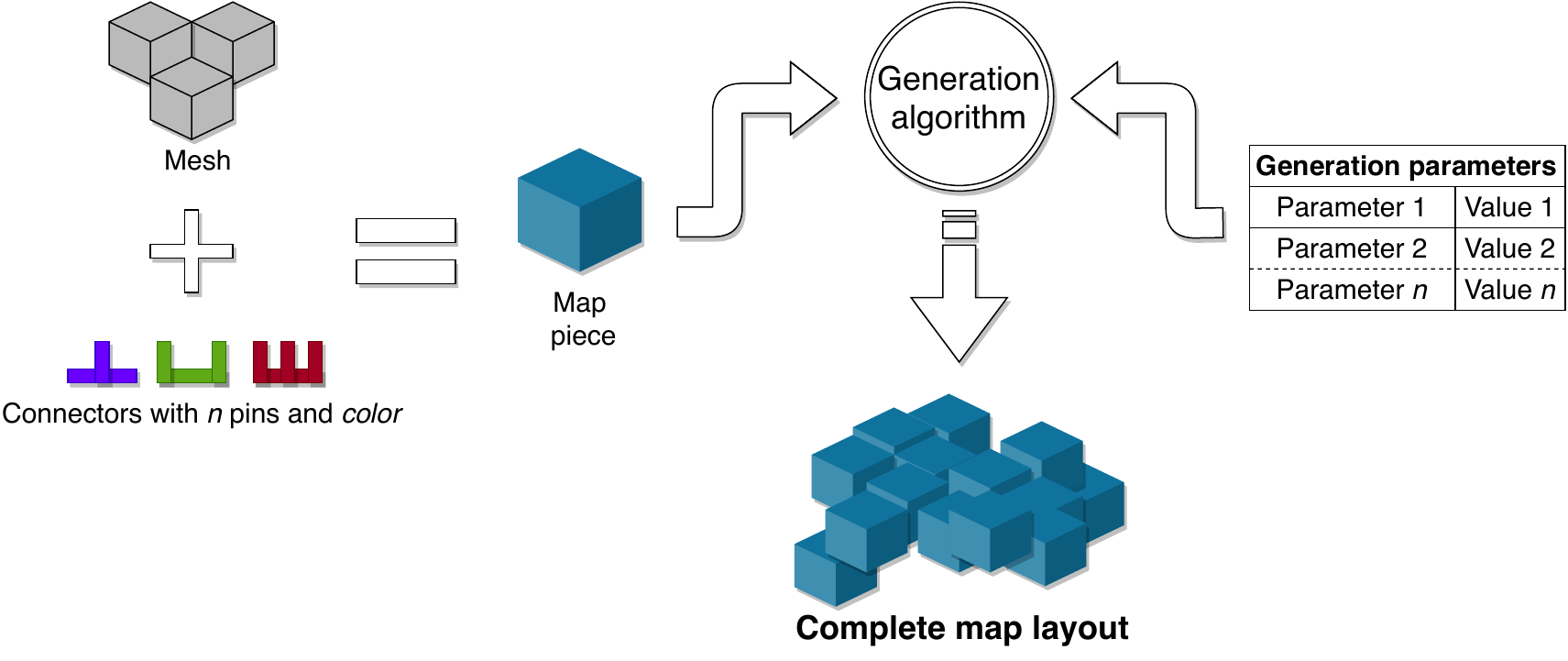}
  \caption{Summary of the snappable meshes map generation process.}
  \label{fig:methodsummary}
\end{figure}

The snappable meshes PCG technique is presented in this section in a game
engine-independent fashion. The technique, summarized in Fig.~\ref{fig:methodsummary},
requires a set of premade \emph{map pieces} to generate maps. These pieces can be
manually created by the game designer and should contain one or more \textit{connectors}.
Connectors, which have a \emph{color} and one or more \emph{pins}, are placed on
the mesh in locations where pieces can snap together. Map pieces, connectors
and pins/colors are detailed in Subsection~\ref{sec:methods:pieces}. The set of
human-designed map pieces, together with a number of generation parameters---also
defined by the game designer---are fed to the \emph{generation algorithm}.
The algorithm is then able to create a playable map, as described in
Subsection~\ref{sec:methods:algorithm}. Certain aspects of the algorithm depend
on the chosen \emph{selection method}. A number of selection methods, and
their influence on the generation algorithm, are presented in
Subsection~\ref{sec:methods:selmethods}. Two metrics for validating the generated
maps are discussed in Subsection~\ref{sec:methods:valid}. Finally, in
Subsection~\ref{sec:methods:class}, the snappable meshes PCG technique is
classified according to the PCG taxonomy proposed by Togelius et al.
\cite{togelius2016intro}.

\subsection{Map Pieces and Connectors}
\label{sec:methods:pieces}

\begin{figure}
    \centering

    \subfloat[\label{fig:piecesandconns:persp}]{
        \includegraphics[width=0.33\linewidth,align=c]{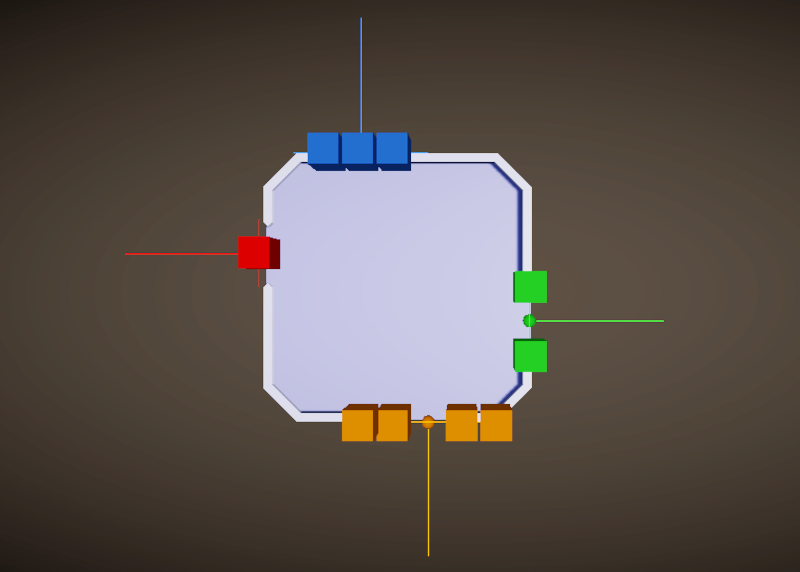}}
    \subfloat[\label{fig:piecesandconns:isol}]{
        \includegraphics[width=0.33\linewidth,align=c]{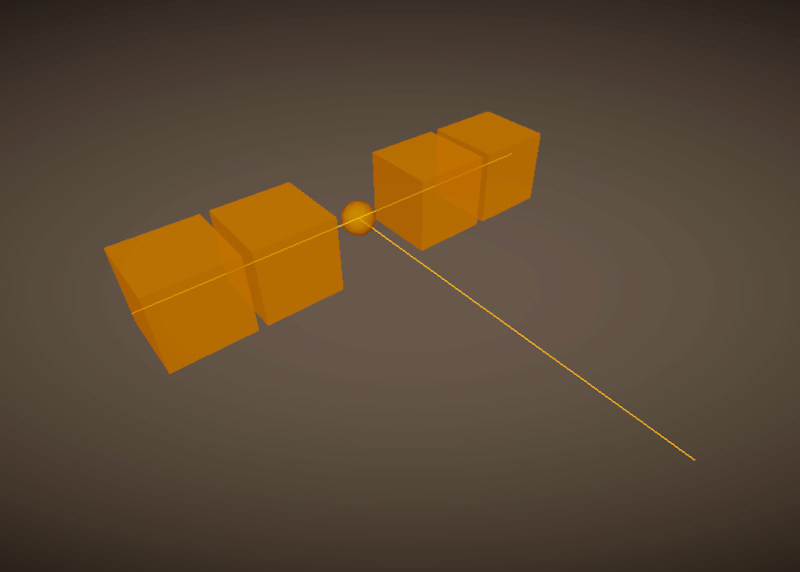}}
    \subfloat[\label{fig:piecesandconns:joined}]{
        \includegraphics[width=0.33\linewidth,align=c]{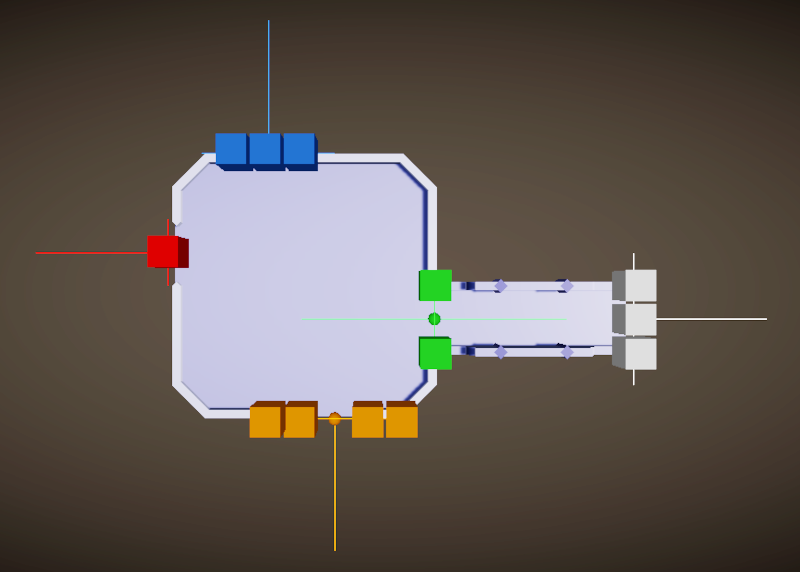}}

    \caption{Map pieces and connectors.
        (a) An example of a map piece with four visible connectors;
        (b) an isolated 4-pin connector with the straight line representing its heading;
        (c) two map pieces snapped together at their compatible connectors.}

    \label{fig:piecesandconns}
\end{figure}

Map pieces are sets of geometry that include one or more connectors. An example of a
map piece with visible connectors is shown in Fig.~\ref{fig:piecesandconns:persp}.
For a map piece to be usable by the generation algorithm it needs to have at least
one associated connector, since these determine where two pieces will be joined, i.e.,
snapped together. Map pieces are in effect content parameters for the generation algorithm.
As shown in Table~\ref{tab:contentparams}, the generation algorithm accepts a list
of pieces for generating the map and an optional list of pieces which can be
used as the \emph{starting piece}, i.e., the first piece to be placed on the map.

\begin{table}[!t]
\caption{Content Parameters for the Generation Algorithm}
\label{tab:contentparams}
\begin{tabular}{p{1.45cm}p{6.55cm}}
\toprule
Parameter & Description\\
\midrule
    \texttt{piecesList}
    & List of map pieces to be used by the algorithm to create the map.\\
    \texttt{useStarter}
    & Boolean. If true, the starting piece will be drawn from
      the \texttt{starterList} instead of the \texttt{piecesList}.\\
    \texttt{starterList}
    & List of map pieces to be used as the starting piece if \texttt{useStarter}
      is true.\\
\bottomrule
\end{tabular}
\end{table}

Compatibility between connectors from different pieces is defined by two
parameters set by the designer: \emph{pin count} and \emph{color}. Connectors
have a heading, represented in Fig.~\ref{fig:piecesandconns:isol} by a straight
line. Pieces with matching connectors can be snapped together, as shown in
Fig.~\ref{fig:piecesandconns:joined}. When two connectors are matched, a copy of
the \emph{tentative piece}---the piece being evaluated for placement on the
map---is positioned so its connector and the connector of the \emph{guide
piece}---a piece previously added to the map---are facing each other. The
connection is made such that the connector pair completely fits/overlaps, unless
a predefined connector distance is set.
As will be discussed next, this connection can be optionally invalidated if the
tentative piece's geometry overlaps with existing map geometry. Matching rules
between connectors are defined as follows:

\begin{enumerate}
    \item Connectors may be compatible if their pin count is equal, or if their
        difference is within a tolerance level defined in the generation parameters.
    \item Connectors may be compatible if their colors match according to a color
        matrix given as a generation parameter. The use of a non-symmetric
        color matrix allows the designer to specify one- or two-way compatibility
        between the connectors of the guide and tentative pieces.
\end{enumerate}

These matching rules are optional. The designer can activate both rules, only
one of them, or even none. If both rules are active, connector compatibility is
established only if both rules are verified. If none of the rules is set, all
connectors in different pieces are compatible and pieces can snap together on
their connectors without restriction. Table~\ref{tab:connparams} summarizes
how the matching rules are given as generation parameters.

\begin{table}[!t]
  \caption{Connection Parameters for the Map Pieces}
  \label{tab:connparams}
  \begin{tabular}{p{1.65cm}p{6.35cm}}
  \toprule
    Parameter & Description \\
  \midrule
    \texttt{matchingRules}
    & Selection. Selects whether to use \textit{none}, \textit{pins}, \textit{colors}
      or \textit{both} as connector matching rules. \\
    \texttt{pinTolerance}
    & Integer. The maximum allowed difference between pin counts in
      two connectors to allow them to be paired up.\\
    \texttt{colorMatrix} & Boolean matrix. Valid color combinations, one-way or
      two-way (i.e., symmetric).\\
  \bottomrule
  \end{tabular}
\end{table}

\subsection{Generation Algorithm}
\label{sec:methods:algorithm}

The pseudo-code of the generation algorithm is presented in Algorithm~\ref{alg:arena_gen},
and the general algorithm parameters are described in Table~\ref{tab:generalparams}.
The algorithm begins by selecting the starting piece (line~\ref{alg:sel_start}) and
placing it on the map (line~\ref{alg:start_place}). The exact process of selecting the
starting piece depends on the chosen selection method, and is detailed in the next
subsection. By default, the selection method should choose the starting piece from the
\texttt{pieceList}, i.e., from the list of pieces used during the generation process.
However, if the \texttt{useStarter} option is enabled, the selection method will
instead pick the starting piece from a separate \texttt{starterList}, as described in
Table~\ref{tab:contentparams}. In any case, when a
piece is selected for placement, it is not used directly. Instead, a copy is made
and it is the copy that is placed on the map. Thus, pieces in these lists act
as blueprints for the pieces actually deployed during the generation process.

\begin{figure}[!t]
\removelatexerror

\begin{algorithm}[H]
  \DontPrintSemicolon
  \caption{Map generation.\label{alg:arena_gen}}

  \SetKwRepeat{Do}{do}{while}
  \SetKw{KwIs}{is}
  \SetKw{KwNot}{not}
  \SetKw{KwAnd}{and}
  \SetKw{KwTerSel}{?}
  \SetKw{KwTerOth}{:}

  \SetKwData{I}{i}
  \SetKwData{StartingPiece}{startingPiece}
  \SetKwData{GuidePiece}{guidePiece}
  \SetKwData{NewPiece}{newPiece}
  \SetKwData{TentativePiece}{tentativePiece}
  \SetKwData{SelMethod}{selMethod}
  \SetKwData{MaxPieces}{maxPieces}
  \SetKwData{PiecesList}{piecesList}
  \SetKwData{UseStarter}{useStarter}
  \SetKwData{StarterList}{starterList}
  \SetKwData{MaxFails}{maxFails}
  \SetKwData{FailCount}{failCount}
  \SetKwData{Connection}{connection}
  \SetKwData{ConnList}{connList}
  \SetKwData{Map}{map}
  \SetKwData{CheckOverlaps}{checkOverlaps}

  \SetKwData{SelectStartingPiece}{SelectStartingPiece}
  \SetKwData{SelectGuidePiece}{SelectGuidePiece}
  \SetKwData{GetRandomItem}{GetRandomItem}
  \SetKwData{GetConnectionsWith}{GetConnectionsWith}
  \SetKwData{IsOverlap}{IsOverlap}
  \SetKwData{SnapWith}{SnapWith}
  \SetKwData{Remove}{Remove}
  \SetKwData{InitializeWith}{InitializeWith}

  \StartingPiece $\leftarrow$ \SelMethod.\SelectStartingPiece{\UseStarter \KwTerSel \StarterList \KwTerOth \PiecesList} \label{alg:sel_start} \\
  \Map.\InitializeWith{\StartingPiece} \label{alg:start_place} \\
  \GuidePiece $\leftarrow$ \StartingPiece \label{alg:guide_start}\\
  \Do{\GuidePiece \KwIs \KwNot none}
  {
    \label{alg:loop_main}
    \FailCount $\leftarrow$ 0 \label{alg:failcount_init} \\

    \Connection $\leftarrow$ \textit{none} \label{alg:connection_init} \\

    \Do{\Connection \KwIs none \KwAnd \FailCount $<$ \MaxFails}
    {

      \label{alg:loop_seltent1}

      \TentativePiece $\leftarrow$ \PiecesList.\GetRandomItem{} \label{alg:tent_getrnd} \\
      \ConnList $\leftarrow$ \GuidePiece.\GetConnectionsWith{\TentativePiece} \label{connlst_get} \\

      \While{\ConnList \KwIs \KwNot empty \KwAnd \Connection \KwIs none}
      {
        \label{alg:loop_selconn1}

        \Connection $\leftarrow$ \ConnList.\GetRandomItem{} \label{alg:conn_get}

        \If{\CheckOverlaps \KwAnd \Map.\IsOverlap(\Connection)   \label{alg:check_overlap}}
        {
          \ConnList.\Remove(\Connection)  \label{alg:conn_remove}

          \Connection $\leftarrow$ \textit{none}  \label{alg:loop_selconn2}
        }
      }
      \If{\Connection \KwIs \KwNot none \label{alg:is_valid_pair}}
      {
        \GuidePiece.\SnapWith{\TentativePiece, \Connection} \label{alg:snap}\\
      }
      \Else
      {
        \FailCount $\leftarrow$ \FailCount + 1
      }
    } \label{alg:loop_seltent2}
    \GuidePiece $\leftarrow$ \SelMethod.\SelectGuidePiece{\Map} \label{alg:guide_next}
  } \label{alg:theend}

\end{algorithm}

\end{figure}

Before entering the main loop of the algorithm, the starting piece is selected as the guide
piece (line~\ref{alg:guide_start}), since it is the only piece currently placed on the map.
When the main loop begins (line~\ref{alg:loop_main}), the \texttt{failCount}
and \texttt{connection} variables are initialized to zero and \textit{none}, respectively
(lines~\ref{alg:failcount_init}--\ref{alg:connection_init}). The former counts how many
failed connection attempts have occurred between the current guide piece and tentative
pieces. The latter represents a valid connection pair between the current guide and
tentative pieces.

The algorithm then enters the tentative piece selection and placement loop
(lines~\ref{alg:loop_seltent1}--\ref{alg:loop_seltent2}). A tentative piece is randomly
selected from \texttt{pieceList} (line~\ref{alg:tent_getrnd}), and all possible connector
pairings between the guide piece and the tentative piece are evaluated. Valid pairings
are stored in a temporary list (line~\ref{connlst_get}). A connector pairing is considered
valid if, and only if, the following conditions are met:

\begin{enumerate}
  \item Both connectors are unused, thus available for pairing.
  \item Both connectors fulfill the criteria described in the previous
    subsection, in accordance with the \texttt{matchingRules} parameter
    (see Table~\ref{tab:connparams}).
\end{enumerate}

The next steps---i.e., the while loop in lines~\ref{alg:loop_selconn1}--\ref{alg:loop_selconn2}---depend
on whether overlap verification (i.e., the \texttt{checkOverlaps} parameter, see
Table~\ref{tab:generalparams}) is enabled or not. If this verification is disabled, the while
loop finishes right after a random pairing is drawn from the temporary list and placed in
the \texttt{connection} variable (line~\ref{alg:conn_get}). However, if overlap verification
is enabled, the algorithm will check if the tentative piece, when connected to the guide piece
by the randomly selected pairing at a distance of \texttt{pieceDistance} (see
Table~\ref{tab:generalparams}), overlaps with existing geometry (line~\ref{alg:check_overlap}).
If so, the pairing is removed from the temporary list (line~\ref{alg:conn_remove}) and the
\texttt{connection} variable is set to \textit{none} (line~\ref{alg:loop_selconn2}). In such
case, the while loop (lines~\ref{alg:loop_selconn1}--\ref{alg:loop_selconn2}) continues until
a non-overlapping solution is found or there are no more pairings left in the temporary list.

\begin{table}[!t]
  \caption{General Parameters for the Generation Algorithm}
  \label{tab:generalparams}
  \begin{tabular}{p{1.65cm}p{6.35cm}}
  \toprule
    Parameter & Description \\
  \midrule
    \texttt{maxFails}
    & Integer. Number of tentative pieces that the algorithm tries to connect with a guide
        piece.\\
    \texttt{pieceDistance}
    & Real value. The spacing between two joined connectors.\\
    \texttt{checkOverlaps}
    & Boolean. If true, connections will be invalidated if there is an overlap
        between tentative piece and map geometry.\\
  \bottomrule
  \end{tabular}
\end{table}

If the previous step yielded a valid pairing (line~\ref{alg:is_valid_pair}), the guide and
tentative pieces are finally snapped together at that location (line~\ref{alg:snap}) and with a
distance defined by the \texttt{pieceDistance} parameter (see Table~\ref{tab:generalparams}). However,
if no valid pairing was found, the algorithm will keep the same guide piece and randomly select
another tentative piece from \texttt{pieceList}. This process is repeated until a valid pairing
is found or a limit of failed attempts (defined in the \texttt{maxFails} parameter, see
Table~\ref{tab:generalparams}) is reached for the current guide piece (line~\ref{alg:loop_seltent2}).

Whether or not a new piece was placed on the map during the tentative piece selection and
placement loop (lines~\ref{alg:loop_seltent1}--\ref{alg:loop_seltent2}), a new guide piece will
be chosen by the selection method (line~\ref{alg:guide_next}). As will be discussed in the next
subsection, selection methods decide the guide piece based on the pieces currently placed on
the map. Consequently, if the tentative piece selection and placement loop was unsuccessful,
the selection method will determine the next guide piece based on exactly the same scenario.
Thus, if the selection method is deterministic, the same guide piece will be picked
again. This leads to an infinite loop if no other piece can be connected with the guide piece in
question. Therefore, the algorithm needs to detect if the same guide piece is chosen two times in
a row and the number of free connectors in it remains the same---which most likely means the
generation process is unable to go any further. In the proposed algorithm, this detection
mechanism is assumed to be encapsulated in the \texttt{SelectGuidePiece()} function
(line~\ref{alg:guide_next}), and works by returning \textit{none} when such a scenario is
detected. The function may also return \textit{none} according to the internal logic of the
selection method in place, as discussed in the next subsection. In any case, when no guide
piece is returned, the map generation ends (line~\ref{alg:theend}).

\subsection{Selection Methods}
\label{sec:methods:selmethods}

\begin{figure}[t!]
  \centering
  \mbox{}
  \subfloat[\label{fig:genoutputarena}]{
    \includegraphics[width=0.48\linewidth,align=c]{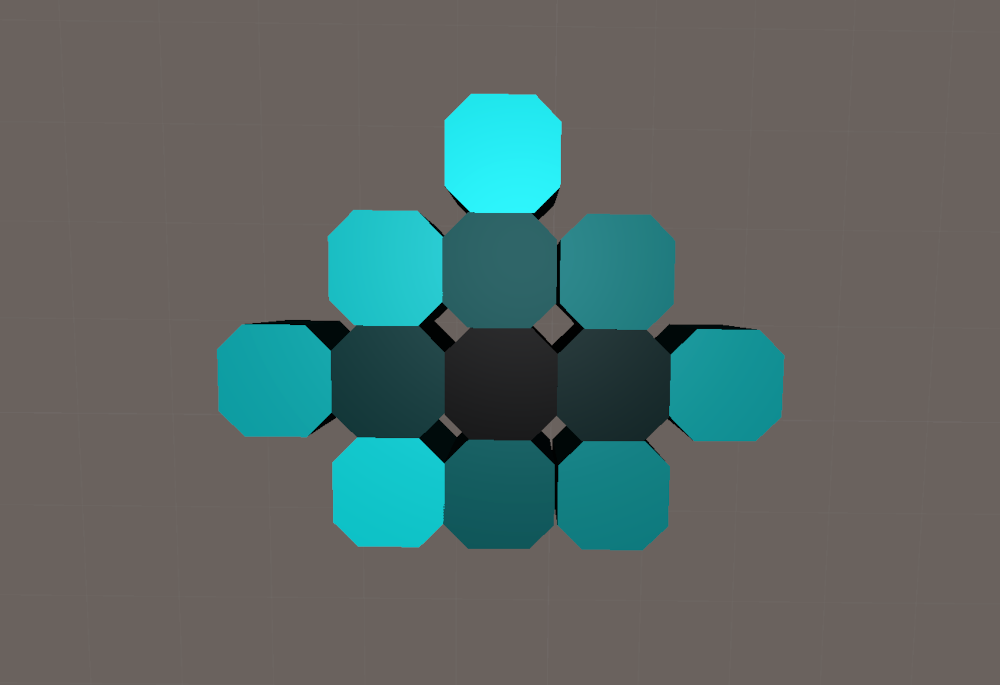}}
  \subfloat[\label{fig:genoutputcorridor}]{
    \includegraphics[width=0.48\linewidth,align=c]{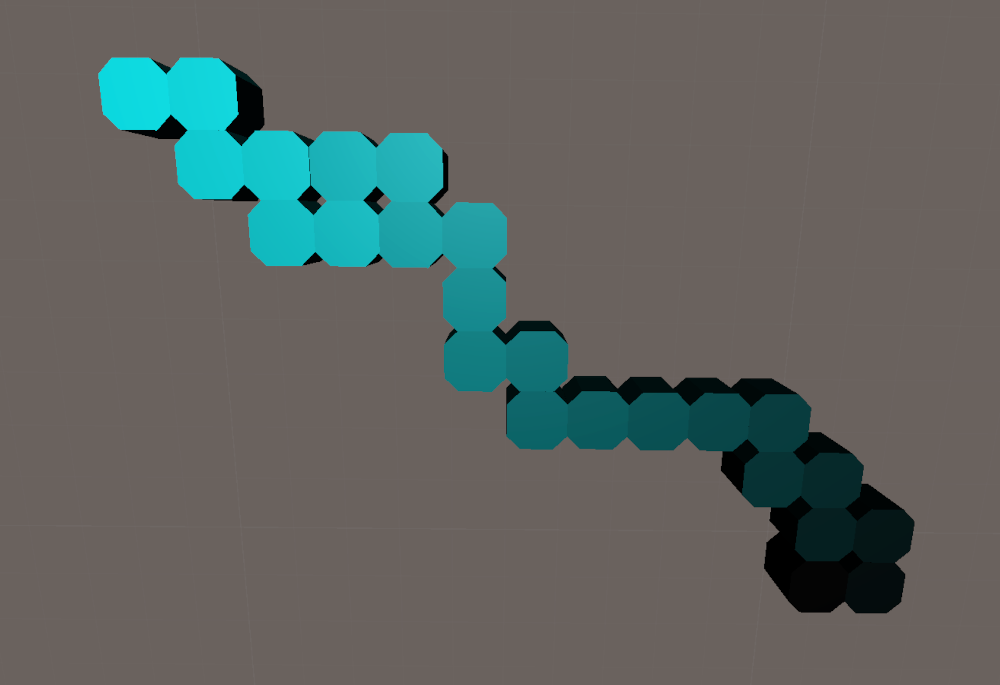}}\\
  \mbox{}
  \subfloat[\label{fig:genoutputstar}]{
    \includegraphics[width=0.48\linewidth,align=c]{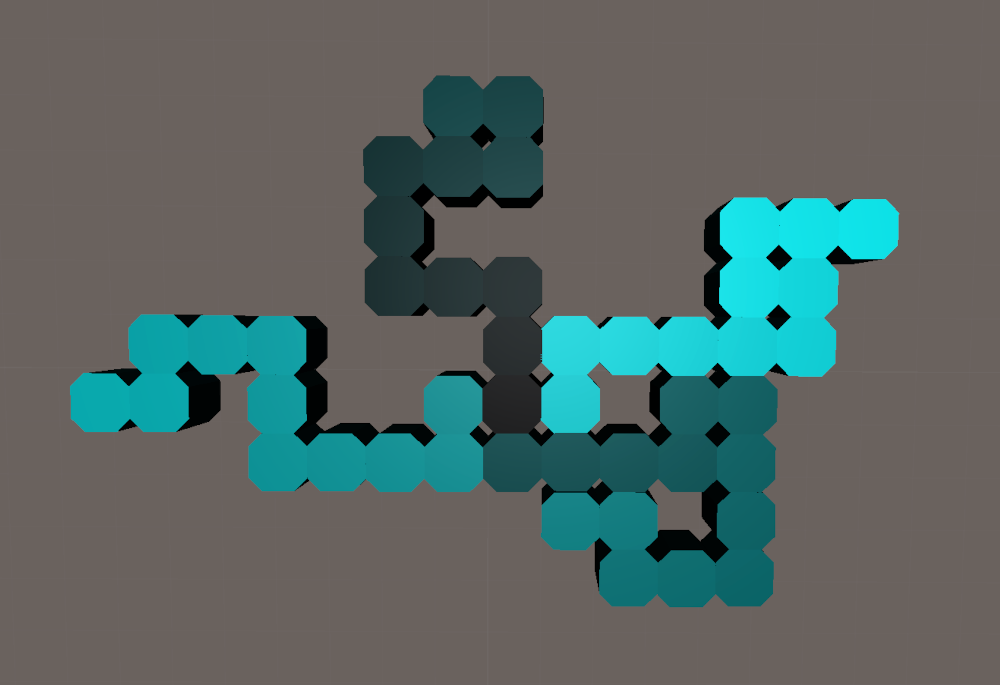}}
  \subfloat[\label{fig:genoutputbranch}]{
    \includegraphics[width=0.48\linewidth,align=c]{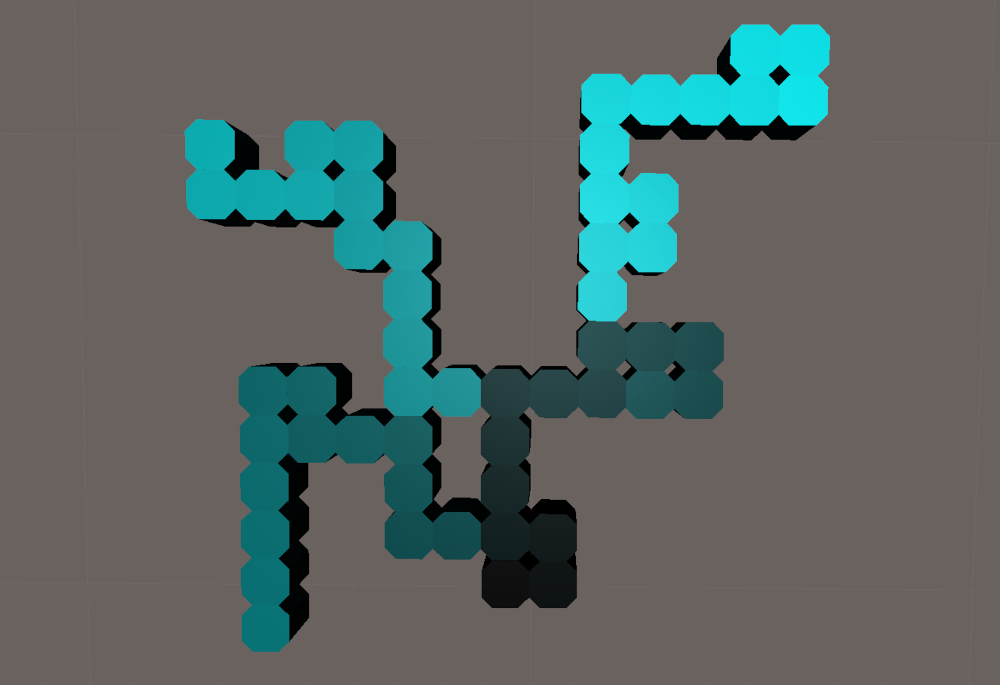}}
  \caption{Example outputs of the proposed selection methods, namely
    (a) \emph{arena} selection method,
    (b) \emph{corridor} selection method,
    (c) \emph{star} selection method,
    and,
    (d) \emph{branch} selection method.
    In these examples, the piece color
    is lighter the later it is placed on the map. Thus, the starting piece is shown with the
    darkest color, while the last snapped piece appears with the lightest color. Pieces are
    very simple and placed in grid-like fashion in order to highlight how the different
    selection methods work.
    }
  \label{fig:genoutput}
\end{figure}

The different selection methods produce distinct map layouts by determining how the generation
algorithm selects the starting piece (line~\ref{alg:sel_start} of Algorithm~\ref{alg:arena_gen})
and the guide pieces (line~\ref{alg:guide_next} of Algorithm~\ref{alg:arena_gen}). Four methods
are proposed and discussed here, namely \emph{arena}, \emph{corridor}, \emph{star} and
\emph{branch}.
Fig.~\ref{fig:genoutput} shows examples of maps created by each of these methods.

Different methods have specific parameter sets, as detailed in Table~\ref{tab:smparams}.
The first parameter, \texttt{starterConTol}, impacts all methods, determining the connector
tolerance for selecting the starting piece. Selection methods choose the starting piece
based on the amount of connectors it has, and \texttt{starterConTol} provides a tolerance
in this selection.
Subsections~\ref{sec:methods:selmethods:arena}--\ref{sec:methods:selmethods:branch}
describe each of the four methods and their parameter sets with additional detail.

\begin{table*}[!t]
  \caption{Parameters for the Selection Methods}
  \label{tab:smparams}
  \begin{tabular}{llp{12.7cm}}
  \toprule
    Method & Parameter & Description \\
  \midrule
      All
      & \texttt{starterConTol}
        & Integer. The starting piece is selected among the set of pieces with most
          (\textit{arena}, \textit{star}) or fewest (\textit{branch}, \textit{corridor})
          connectors, $n_\text{max}$ or $n_\text{min}$, respectively. This parameter
          represents the tolerance, in number of connectors, from the piece(s) with
          most (or fewest) connectors, allowing pieces with as few as
          $n_\text{max}-\texttt{starterConTol}$ (or as much as
          $n_\text{min}+\texttt{starterConTol}$) connectors to be selected
          as the starting piece.\\
  \hdashline[.5pt/1pt]
    \emph{Arena}, \emph{Corridor}
    & \texttt{maxPieces}
      & Integer. The maximum number of pieces the generator will place on the map.\\
  \hdashline[.5pt/1pt]
      \emph{Star}
      & \texttt{armLength}
        & Integer. The average amount of pieces an arm of the ``star'' will have.\\
      & \texttt{armLengthVar}
        & Integer. The maximum variation from \texttt{armLength} in each arm.\\
  \hdashline[.5pt/1pt]
      \emph{Branch}
      & \texttt{branchCount}
        & Integer. The number of branches to be created. \\
      & \texttt{branchLength}
        & Integer. The average amount of pieces a branch will have.\\
      & \texttt{branchLengthVar}
        & Integer. The maximum variation from \texttt{branchLength} in each branch.\\
  \bottomrule
  \end{tabular}
\end{table*}

\subsubsection{The arena selection method}
\label{sec:methods:selmethods:arena}

This method (example in Fig.~\ref{fig:genoutputarena}) aims to create maps that sprawl
in all directions from the starting piece, covering the surrounding area with geometry.
It is a generic and useful approach, having been used for the Trinity game with
satisfactory results, as discussed in Section~\ref{sec:casestudy:trinity}.

The piece with most connectors is selected as the starting piece. If there are multiple
pieces with the same highest number of connectors and/or if more pieces are considered
for selection due to the \texttt{starterConTol} parameter, one of them is picked at random.
The goal is to have a central piece with many connections so that geometry can spread in
various directions.

For the selection of the next guide piece, the \emph{arena} method checks if the current
guide piece has unused connectors, and if so, keeps it as the guide piece in the next
tentative piece selection and placement loop. Otherwise, the piece placed immediately
after the current guide piece is selected as the next guide piece. This way, by
performing a breadth-first traversal of the search space, the method preferentially
fills center locations near the original starting piece. The \emph{arena} method
stops when the number of pieces placed on the map is higher than \texttt{maxPieces}, as
described in Table~\ref{tab:smparams}.

\subsubsection{The corridor selection method}
\label{sec:methods:selmethods:corridor}

The goal of the \emph{corridor} selection method is to generate narrow maps where the
geometry seemingly follows a line, as shown in Fig.~\ref{fig:genoutputcorridor}.

The piece with least connectors is selected as the starting piece. If there are multiple
pieces with the same lowest number of connectors and/or if more pieces are considered
for selection due to the \texttt{starterConTol} parameter, one of them is selected at
random. The selected guide piece is always the most recently placed piece, such that
pieces will have at most two connections: one with the previous placed piece and the
other with next compatible tentative piece. Thus, the method performs a depth-first
traversal of the search space.

The main motivation for implementing the \emph{corridor} method is its potential
suitability for ``Capture the Flag''-style matches, in which two players or teams must
capture the enemy's flag from their end of the map and bring it back.

\subsubsection{The star selection method}
\label{sec:methods:selmethods:star}

This method is a mix of the \emph{arena} and \emph{corridor} selection methods,
creating lanes sprawling from the starting piece and ending when that piece has no
more available connectors. Thus, the number of arms in the ``star'' is equal to the
number of available connectors in the starting piece, which acts as a central hub for
the map layout.

The starting piece is selected the same way as in the case of the
\emph{arena} method, i.e., based on the highest amount of connectors. As with the
other methods, the designer can force a specific piece or pieces to be used as the
central hub by manipulating the \texttt{useStarter} and \texttt{starterList} parameters
(see Table~\ref{tab:contentparams}).

Each arm of the star, $i$, will have a uniformly random length (i.e., number of pieces)
of $l_i=\text{\texttt{armLength}} \pm \text{\texttt{armLengthVar}}$ (see
Table~\ref{tab:smparams}), which, nonetheless, cannot be smaller than one. For this
purpose, the selected guide piece will be the last placed piece while the length of
the current arm $i$ is lower than $l_i$. When the length of the current arm reaches
$l_i$, the starting piece (acting as the central hub) is returned again as the guide
piece, allowing for the creation of a new arm. If this piece has no connectors left,
the method returns \textit{none}, ending the generation process.

The \emph{star} selection method will try its best to generate a map with (a) a
number of arms equal to the amount of connectors in the starting piece, and, (b)
arms with length within the interval $\text{\texttt{armLength}} \pm
\text{\texttt{armLengthVar}}$. However, this might not be possible for two reasons.
First, the generation algorithm may fail to find a tentative piece compatible with
the guide piece within \texttt{maxFails} attempts (see Table~\ref{tab:generalparams}).
Second, if there are pieces with a single connector, these would effectively work as
premature arm terminals, in which case a new arm should be opened.
Fig.~\ref{fig:genoutputstar} shows a map generated with the \emph{star} selection
method with four arms and arm length between 10 and 13 pieces.

A possible use case for the \textit{star} selection method are ``King of the Hill''-type
game modes, in which players converge onto the middle of the map, fighting for control.

\subsubsection{The branch selection method}
\label{sec:methods:selmethods:branch}

This method (example in Fig.~\ref{fig:genoutputbranch}) creates branches in a similar manner
to the \emph{star} method, except that it does not return the
starting piece as the guide piece when a branch is finished. Instead, the \emph{branch}
selection method selects one of the previously placed pieces to start a new branch,
repeating this until the specified number of branches (\texttt{branchCount},
Table~\ref{tab:smparams}) is reached. Furthermore, in contrast to the \emph{star} method
and similarly to the \emph{corridor} method, pieces with fewer connectors are
preferentially selected as the starting piece.

The exact process of selecting the guide piece is as follows. The first branch originates
at the starting piece. The branch grows by using the last placed piece as the guide piece
until its maximum size (uniformly randomly drawn from the interval $\text{\texttt{branchLength}}
\pm \text{\texttt{branchLengthVar}}$) is reached. At this time, it is necessary to select
another map piece to be the root piece of a new branch. This selection is done by considering
a list of pieces already placed on the map, from the starting piece to the last successfully
placed tentative piece. The root piece is then selected by ``jumping'' from the starting
piece in this list (index 0) to one of the other pieces. The base jump value is given by:

\begin{equation*}
  j_\text{base}=\max\left\{1,\left\lfloor\frac{\texttt{\footnotesize{branchCount}}}{\texttt{\footnotesize{branchLength}}}\right\rfloor\right\}
\end{equation*}

\noindent For each new branch $i$, where $i \in \{0, \dots, \text{\texttt{branchCount}}-1\}$, the
effective jump from the starting piece is given by:

\begin{equation*}
  j_i=i \cdot j_\text{base}
\end{equation*}

\noindent i.e., new branches will be based off pieces that were deployed after the root
piece of the previous branch. If the new guide/root piece does not have available connectors,
and to avoid premature termination of the generation process, the neighborhood
$[j_i-j_\text{base},j_i+j_\text{base}]$ is searched back and forth from its center, until
a piece with available connectors is found and returned as the new guide/branch root piece.
If such piece is not found in the neighborhood, the method returns \textit{none} and the
generation process terminates.

The \textit{branch} method offers a way of generating non-linear maps without a central hub
region, increasing exploration possibilities for players.

\subsection{Automatic Validation of Map Navigability}
\label{sec:methods:valid}

Depending on the generation parameters, or simply due to ``unlucky'' seeds, it is possible
that the procedurally generated maps are mostly untraversable or contain several unconnected
regions. Although the snappable meshes technique produces fully explainable outputs, designers
may be more interested if a generated map is actually playable \cite{zhu2018explainable}.
Therefore, some sort of automatic validation of map navigability becomes an important, if
not crucial aspect in this type of algorithm \cite{lelis2017procedural}. We propose two
metrics for this purpose, which can be obtained by deploying a predetermined amount $n$ of
randomly distributed navigation points on the map and verifying their connectivity using
standard path finding. These metrics are:

\begin{enumerate}
    \item The average percentage of valid connections between navigation points, or
        $\overline{c}$.
    \item The relative area of the largest fully-connected (i.e., fully-navigable) region, or
        $A_r^\text{max}$.
\end{enumerate}

The first metric, $\overline{c}$, is given by $\overline{c}=c_t/c_{all}$, where $c_t$ is the
number of traversable connections between all $n$ navigation points, and $c_{all}$ is the
number of all connections, traversable or not. Note that $c_{all}=n(n-1)/2$, i.e., $c_{all}$
is equal to the maximum number of connections between nodes in an undirected graph.
Consequently, this approach has $\mathcal{O}(n^2)$ complexity, which limits the amount of
deployable navigation points, $n$.

The second metric, $A_r^\text{max}$, requires determining the various
fully-navigable---but separate---regions in the generated map. This can be done by analyzing
each pair of connected navigation points as follows: (1) if they are both isolated, create a
new cluster containing them; (2) if one is isolated and the other is not, add the former to
the latter's cluster; or, (3) if they
belong to different clusters, merge the clusters. Then, the approximate relative area
represented by each cluster or region can be obtained by dividing the number of navigation
points it contains by $n$, with $A_r^\text{max}$ representing the largest of these areas. Again,
this procedure has $\mathcal{O}(n^2)$ complexity and the same caveat.

The first metric offers a general view of map navigability. The second metric is
arguably more useful, as it can be used, for example, to select a playable area for
deploying agents and/or in determining if the largest region represents a large enough
area of the map to be usable or above a predefined threshold.

While the algorithmic complexity of these procedures is $\mathcal{O}(n^2)$, in practice---and
as will be discussed in Subsection~\ref{sec:casestudy:benchmarks:validation}---the upper bound
for the number of navigation points, $n$, is well above what is required for performing fast
and accurate computations of the two metrics.

\subsection{Classification}
\label{sec:methods:class}

The snappable meshes technique is a constraint solver with no backtracking, focused
on speed and simplicity. The method is able to quickly create maps---as will be shown
in Section~\ref{sec:casestudy}---and is thus appropriate for runtime generation.
Simplicity is a consequence of the visual and easy-to-understand constraints, mainly
in the form of mesh connector rules.

According to the PCG taxonomy described by Togelius et al. \cite{togelius2016intro},
the proposed technique is \emph{offline}, \emph{necessary}, \emph{generic} and
\emph{stochastic}. It is \emph{offline} since maps are generated during game
development or before the start of a game session. The method is \emph{necessary}
because it provides the main structure of the levels (i.e., the 3D maps). Content
is generated without taking the player's previous behavior into account,
hence snappable meshes is a \emph{generic} (or \textit{experience-agnostic}
\cite{yannakakis2018artificial}) PCG technique, as opposed to an \textit{adaptive}
(or \textit{experience-driven} \cite{yannakakis2018artificial}) one. Finally, it is
a \textit{stochastic} technique, as it offers considerable map variability given the
same set of input parameters\footnote{The method is technically \textit{deterministic},
in the sense that it will generate the same output if also given the same seed.}.

There are three other categories in this taxonomy where the snappable meshes technique
does not fall under a specific classification. The first concerns autonomy, which
differentiates between \textit{autonomous} versus \textit{MICC} PCG approaches. As
stated in Section~\ref{sec:background}, this separation is not clear-cut. On one hand,
the technique works autonomously after the designer defines its input (parameters and
map pieces).
On the other, it is the designer who creates the fundamental building blocks of the
generated maps, hence affecting their look and feel.
Thus, the snappable meshes technique is essentially autonomous,
though stating that it does not employ some form of ``mixed-authorship'' seems
inaccurate.
The second category, \textit{degree and dimensions of control} (or \textit{controllabiliy}
\cite{yannakakis2018artificial}), specifies the ways in which content generation can be
controlled, for example using random seeds or content/parameter vectors. The proposed
technique generates content using both approaches simultaneously. More specifically, an
(optionally seeded) pseudo-random number generator can drive content creation based on the
content and parameters specified by the designer.
The third category, \textit{constructive versus generate-and-test}, defines whether content is
generated in one pass (\textit{constructive}) or oscillates between content generation and
testing in a loop, until a suitable output is found. The snappable meshes methodology allows
for both approaches. It has been used as a \textit{constructive} method in the Trinity game,
as will be discussed in Subsection~\ref{sec:casestudy:trinity}, but supports a
\textit{generate-and-test} loop using validation metrics such as the ones presented in the
previous subsection.

\section{A Prototype Implementation in Unity}
\label{sec:impl}

A standalone demonstration prototype of the snappable meshes PCG technique was implemented in
the Unity game engine, leveraging its editor tools to handle the input of the human designer.
This is a simple reference implementation based on the code originally developed for the
\textit{Trinity} game, discussed in the next section, and is provided as a Unity project---i.e.,
it must be experimented in the Unity editor. The aim of this prototype is to allow designers
and researchers to explore the proposed technique, particularly in how the chosen set of map
pieces and the different algorithm parameters influence map generation. The prototype is
bundled with two predefined scenes, allowing interested users to get started quickly. These
scenes, denoted \textit{Benchmark} and \textit{Artistic}, are configured to use contrasting
sets of map pieces, and a number of example maps, further discussed in
Section~\ref{sec:casestudy}, are respectively shown in Fig.~\ref{fig:cases} and
Fig.~\ref{fig:altmaps}. The code is fully documented and available at
\url{https://github.com/VideojogosLusofona/snappable-meshes-pcg} under the open-source
Apache 2.0 license, meaning it can be freely adapted and used in commercial contexts.

In Subsection~\ref{sec:impl:workflow}, the reference implementation's designer workflow and
its editor-based user interface are described, while relevant implementation details are
highlighted in Subsection~\ref{sec:impl:details}.

\subsection{Designer Workflow and User Interface}
\label{sec:impl:workflow}

\begin{figure}[!t]
  \centering
  \includegraphics[width=1\linewidth]{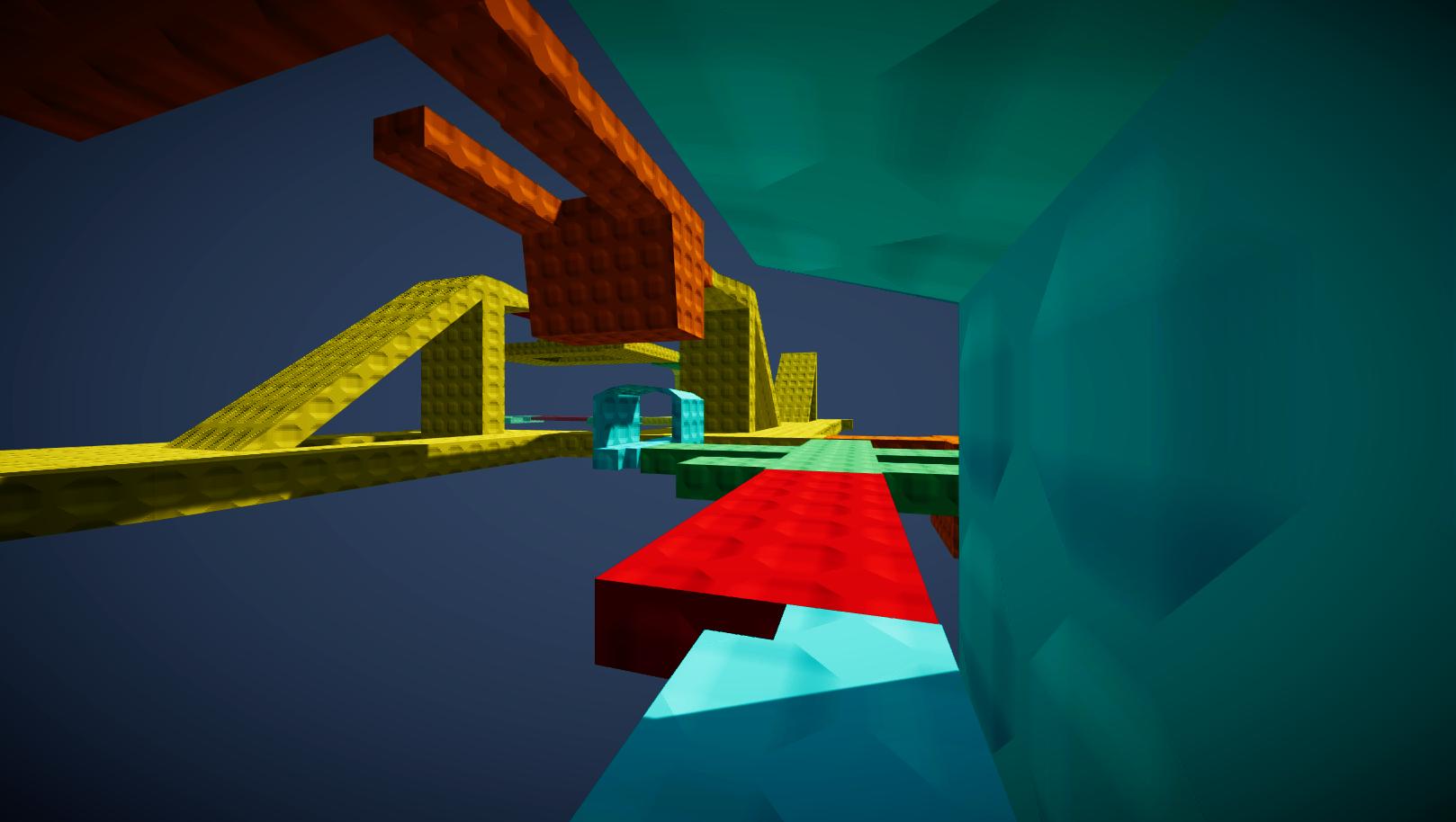}
  \caption{Screenshot of the first-person demo mode on a map generated in the \textit{Benchmark}
    scene.}
  \label{fig:navdemo}
\end{figure}

The designer workflow is divided into roughly three parts:

\begin{enumerate}
    \item Setup and configuration of the generation process. At this stage, the designer imports
        and/or selects the map pieces to use for the generation process, and defines the generation
        parameters.
    \item Map generation and validation assessment. The designer generates the map, which
        appears in the scene view; the validation metrics---described in
        Subsection~\ref{sec:methods:valid}---appear in the console, allowing the designer
        to quickly assess the map's navigability.
    \item Demo of an NPC traversing the map. The two previous steps occur in editor mode,
        i.e., when the ``game'' is not running. To get a first-person feeling for the
        generated map, the designer can enter play mode, in which case a demo of an NPC
        traversing the map starts, as exemplified in Fig.~\ref{fig:navdemo}.
\end{enumerate}

Naturally, this workflow can be repeated and iterated upon until the designer is satisfied
with the results. The next subsections further detail these steps, as well as the user
interfaces which allow for this interaction.

\subsubsection{Setup and configuration of the generation process}
\label{sec:impl:workflow:setup}

\begin{figure*}[!t]
  \centering
  \begingroup
  \setlength{\tabcolsep}{1.5pt}
  \begin{tabular}{ccc}

   \parbox[c][][t]{0.326\textwidth}{
   \subfloat[\label{fig:params:buttons}]{
    \includegraphics[width=1\linewidth]{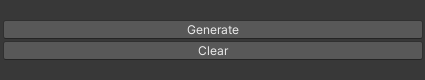}}\\
   \subfloat[\label{fig:params:content}]{
    \includegraphics[width=1\linewidth]{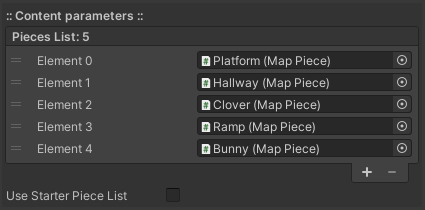}}
   }
   &
   \parbox[c][][t]{0.326\textwidth}{
   \subfloat[\label{fig:params:connection}]{
    \includegraphics[width=1\linewidth]{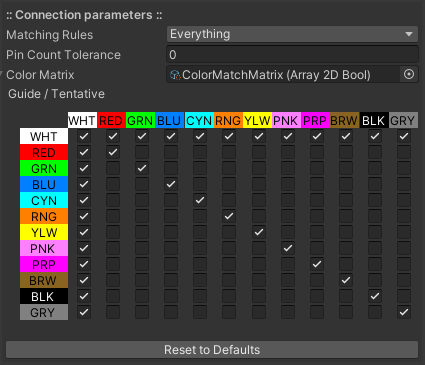}}
   }
   &
   \parbox[c][][t]{0.326\textwidth}{
   \subfloat[\label{fig:params:general}]{
    \includegraphics[width=1\linewidth]{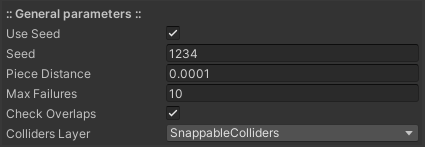}}\\
   \subfloat[\label{fig:params:gm}]{
    \includegraphics[width=1\linewidth]{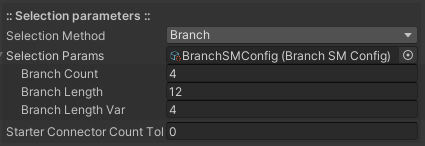}}
   }
  \end{tabular}
 \caption{The map generation parameters and control buttons, part of the
 \texttt{GenerationManager} component, as shown in Unity's inspector. The different
 generation parameter blocks, (b)--(e), essentially correspond to the parameters
 respectively presented in Tables~\ref{tab:contentparams}--\ref{tab:smparams}.
 More specifically, subfigures show:
 (a) the control buttons;
 (b) the content parameters;
 (c) the connection parameters;
 (d) the general parameters;
 and,
 (e) the selection method parameters (here exemplified for the case of the \textit{branch} selection method).}
 \label{fig:params}
 \endgroup
\end{figure*}

The designer sets up the generation process by interacting with the \texttt{GenerationManager}
component, displayed in Unity's inspector---see Fig.~\ref{fig:params}---when the game object of
the same name is selected in the scene's object hierarchy. Here, the designer can generate a new
map or clear an existing one (Fig.~\ref{fig:params:buttons}), select the pieces used to create
the map (Fig.~\ref{fig:params:content}, Table~\ref{tab:contentparams}), define the connection
criteria (Fig.~\ref{fig:params:connection}, Table~\ref{tab:connparams}), specify general
parameters (Fig.~\ref{fig:params:general}, Table~\ref{tab:generalparams}), and choose and
configure a selection method (Fig.~\ref{fig:params:gm}, Table~\ref{tab:smparams}).

Configuring map validation is also performed at this stage. This is done in a separate inspector
panel, which appears when the \texttt{NavController} game object is selected in the object
hierarchy. Among other aspects, the designer can define the dimension of NPCs and/or players
navigating the map, as well as the number of navigation points used for determining the
navigation metrics, as explained in Subsection~\ref{sec:methods:valid}.

\subsubsection{Map generation and validation assessment}
\label{sec:impl:workflow:genvalid}

After configuring the generation process, the designer can create a new map by clicking the
``Generate'' button (Fig.~\ref{fig:params:buttons}). Besides the produced map, the process
also outputs a detailed generation log in Unity's console, listing each selected guide and
tentative pieces, as well as successful and unsuccessful snaps, allowing the designer to
follow all the steps of the generation process.

After the map is generated, the validation step is triggered, and a predefined number of
navigation points are randomly placed on the map's surface in order to obtain the validation
metrics previously described, namely, $\overline{c}$, the map's average relative connectivity,
and, $A_r^\text{max}$, the relative area of the largest fully-connected region. This information
is also provided via a log in Unity's console (separate from the generation log).

These metrics can be used to request a regeneration of the level if the average connectivity
is below a certain threshold and/or the largest region does not represent a large enough
relative area of the generated map. Additionally, $A_r^\text{max}$ can be used to define
the playable area of the map, where new players or NPCs are spawned, for example.

\subsubsection{Demo of an NPC traversing the map}
\label{sec:impl:workflow:demo}

Having successfully generated a map, the designer can enter Unity's play mode, where an NPC
traverses the map using the navigation points (initially created for map validation) as
randomly selected waypoints. The NPC is placed in the largest fully-navigable area of the map,
also determined during the validation procedure, thus avoiding getting stuck in poorly
connected regions. While this demo is not useful in a real-world automatic generation process,
it constitutes a visual aid for the designer to get a first-person awareness of the generated
map. An example of this demo mode is shown in Fig.~\ref{fig:navdemo}.

\subsection{Implementation Details}
\label{sec:impl:details}

In this section we highlight a number of implementation details in the Unity prototype
which might be relevant for helping researchers understand the code and/or for developers
implementing their own versions of snappable meshes.

\subsubsection{Piece Design and Deployment}
\label{sec:impl:details:pieces}

When the designer creates a map piece, it is necessary to specify its mesh, individual
connectors (each with a color and a number of pins), and one or more colliders\footnote{A
collider is Unity's terminology for a bounding volume.} when piece overlap is to be
avoided (as discussed in Subsection~\ref{sec:impl:details:overlaps}). In this context,
human-designed pieces are created as \textit{prefabs}, Unity's implementation of the Prototype
design pattern \cite{nystrom2014game}. Therefore, pieces in the \texttt{piecesList} and
\texttt{starterList} (Table~\ref{tab:contentparams}) are prototypes, and the pieces actually
placed on the map are effectively copies of the original designs.

In this implementation, and for the purpose of snapping two meshes together, tentative pieces
can only be rotated about their vertical axis, so that the involved connectors face each other.
Therefore, meshes will never tilt or flip.

\subsubsection{Overlap Detection}
\label{sec:impl:details:overlaps}

An important part of the proposed technique is the ability to generate maps without
intersecting geometry. To guarantee this, when a tentative piece is selected, an
optional verification ensures that the piece does not overlap with existing
geometry for each of its possible connections (line~\ref{alg:check_overlap} of
Algorithm~\ref{alg:arena_gen}). The \texttt{checkOverlaps} parameter
(Fig.~\ref{fig:params:general}, Table~\ref{tab:generalparams}), determines whether
this verification is performed or not.

For overlap verification to work, pieces must contain one or more box colliders, i.e.,
rectangular cuboid-shaped bounding volumes. These should approximately mirror the
piece's shape, allowing Unity to detect if the tentative piece overlaps with existing map
geometry in a quick and relatively accurate fashion. We opted for box colliders since in
Unity, due to optimization concerns, general convex mesh colliders are limited to 255
triangles and often display inaccurate behavior. These box colliders are tagged in a
separate object layer, so that the application is able to find them while creating the
map and safely delete them when the generation process is finalized. By default, this
layer is named \textit{SnappableColliders} (Fig.~\ref{fig:params:general}).
Nonetheless, the user can specify another name.

\subsubsection{Selection Methods}
\label{sec:impl:details:gm}

The Strategy design pattern \cite{nystrom2014game} was used to decouple the selection
methods from the generation algorithm itself. What this means is that the different
methods are placed in distinct classes derived from a common base class. Any existing
selection method classes are then ``discovered'' using C\#'s reflection,
and matched with an appropriate configuration object. This object is then used by the
\texttt{GenerationManager} component to present the user the available parameters for
the current selection method, chosen from a drop-down list populated during the
``discovery'' process. Fig.~\ref{fig:params:gm} shows what is presented to the user
when the \textit{branch} method is selected.

This approach simplifies the creation of new selection methods, requiring only two
classes to be implemented: one for the selection method itself, and the other for
configuring it.

\subsubsection{Map Validation and Demo}
\label{sec:impl:details:validation}

Map validation occurs after the generation process, and is performed using Unity's
built-in navigation mesh---or \textit{navmesh}---system. A navmesh is a mesh of convex
polygons that define traversable areas on a map. These polygons can be considered
nodes in a graph, with adjacent polygons forming valid paths, or in graph terminology,
a connection between nodes. Thus, a path finding algorithm such as A$^*$ can be used
to determine if a path exists between any two nodes \cite{millington2019ai}.

Unity's navmesh system allows the runtime creation of navmeshes on existing geometry,
and is used in our reference implementation for this purpose. After a navmesh is
created for a generated map, a predetermined number of navigation points is deployed
in the navmesh. At this stage, Unity's path finding system is used for determining if
there are valid paths between each pair of navigation points. With this information,
the metrics discussed in Subsections~\ref{sec:methods:valid} and
\ref{sec:impl:workflow:genvalid} can be easily computed. Finding traversable paths
for the first-person demo (Subsection~\ref{sec:impl:workflow:demo}) is similarly
straightforward.

To aid in the visual inspection of the generated maps, navigation points placed in
the largest fully-traversable region of the map are rendered in green, while points
deployed in other regions are shown in red (see Fig.~\ref{fig:valid}).

\section{Case Studies}
\label{sec:casestudy}

In this section several case studies are investigated with the purpose of
highlighting the potential of the snappable meshes PCG technique, as well as its
limitations. In Subsection~\ref{sec:casestudy:benchmarks} we start by analyzing how
the different parameters and selection methods influence the maps generated with the
\textit{Benchmark} scene, focusing on eight illustrative examples. These same
examples are then dissected from a benchmarking perspective, both in terms of
generation and validation time, as well as concerning the quality of the proposed
validation metrics.
In Subsection~\ref{sec:casestudy:altpieces}, with the goal of further exploring
the generative capabilities of the snappable meshes technique, we evaluate four maps
generated with the \textit{Artistic} scene, since it provides a set of building blocks
which is completely different than those available in the \textit{Benchmark} scene.
Finally, in Subsection~\ref{sec:casestudy:trinity}, we examine how the proposed method
was used to generate the maps for the \textit{Trinity} third-person shooter game,
enabling the desired gameplay style and replayability.

\subsection{Analysis and Validation of Snappable Meshes Using the \textit{Benchmark} Scene}
\label{sec:casestudy:benchmarks}

\subsubsection{Experimenting With the Different Parameters and Selection Methods}
\label{sec:casestudy:benchmarks:piecesmap}

\begin{figure*}[!t]
  \centering
  \subfloat[\label{fig:pieces:platform}]{
    \includegraphics[height=3.5cm,,align=c]{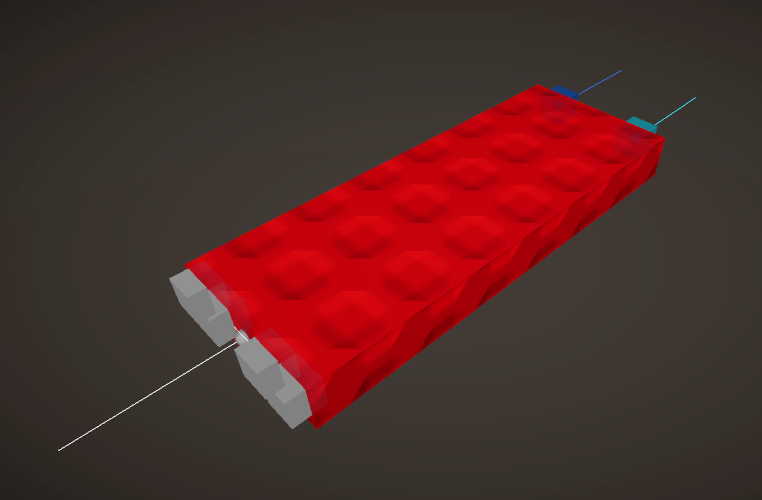}}
  \subfloat[\label{fig:pieces:hallway}]{
    \includegraphics[height=3.5cm,,align=c]{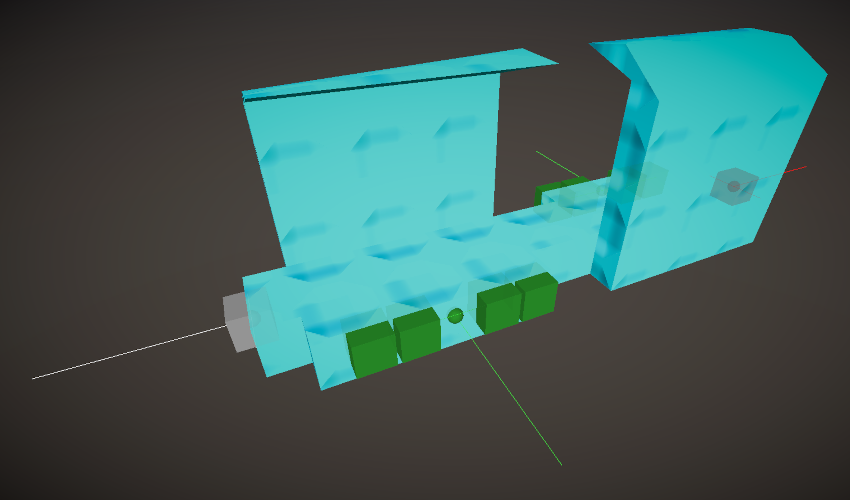}}
  \subfloat[\label{fig:pieces:clover}]{
    \includegraphics[height=3.5cm,align=c]{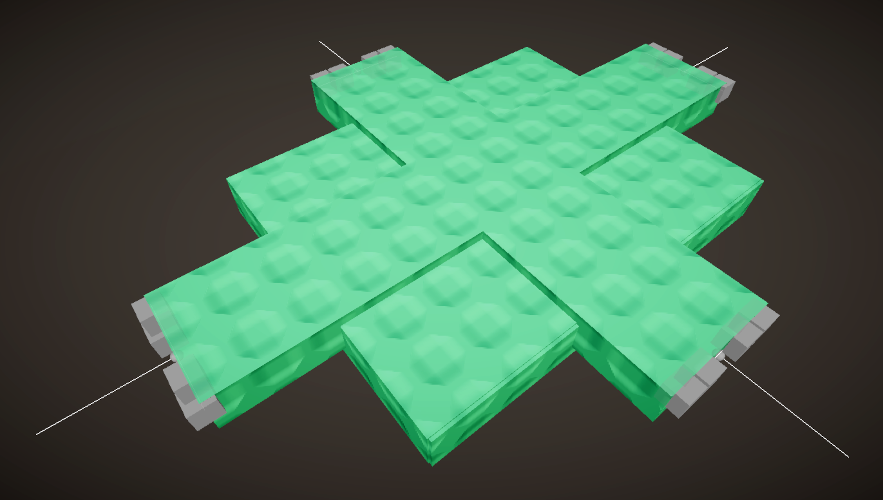}}\\
  \subfloat[\label{fig:pieces:ramp}]{
    \includegraphics[height=3.5cm,,align=c]{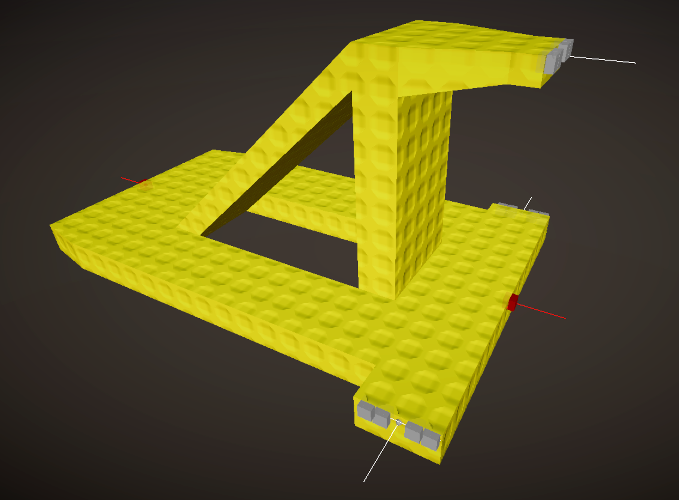}}
  \subfloat[\label{fig:pieces:bunny}]{
    \includegraphics[height=3.5cm,,align=c]{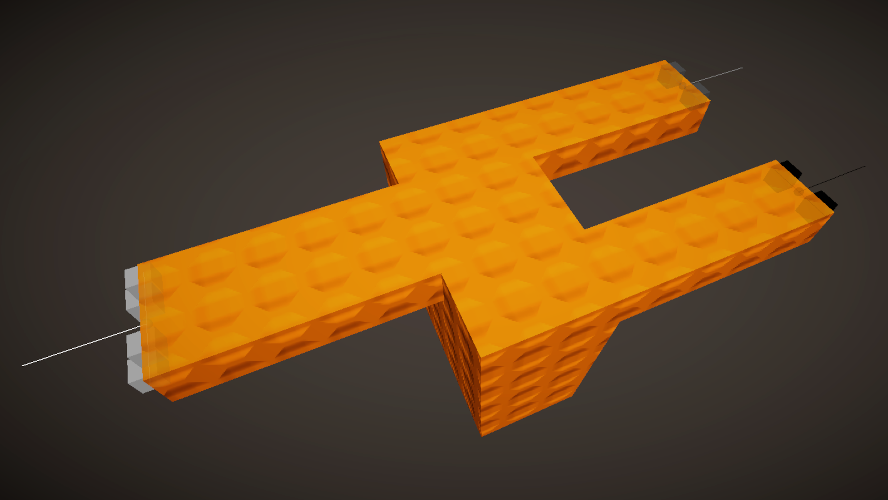}}
  \caption{Map pieces included with the \textit{Benchmark} scene of the Unity
  prototype implementation. Piece names and connector configurations are as follows
  (each box corresponds to a connector with a specific color and pin count):
  (a) ``Platform'': \fcolorbox{black}{cyan}{\texttt{1}}\fcolorbox{black}{blue}{\textcolor{white}{\texttt{1}}}\fbox{\texttt{4}};
  (b) ``Hallway'': \fcolorbox{black}{red}{\textcolor{white}{\texttt{1}}}\fbox{\texttt{1}}\fcolorbox{black}{green}{\texttt{4}}\fcolorbox{black}{green}{\texttt{4}};
  (c) ``Clover'': \fbox{\texttt{4}}\fbox{\texttt{4}}\fbox{\texttt{4}}\fbox{\texttt{4}};
  (d) ``Ramp'': \fcolorbox{black}{red}{\textcolor{white}{\texttt{1}}}\fcolorbox{black}{red}{\textcolor{white}{\texttt{1}}}\fbox{\texttt{4}}\fbox{\texttt{4}}\fbox{\texttt{4}};
  (e) ``Bunny'': \fcolorbox{black}{gray}{\textcolor{white}{\texttt{2}}}\fcolorbox{black}{black}{\textcolor{white}{\texttt{2}}}\fbox{\texttt{4}}.
  Note that white is configured as a wildcard color in the \textit{Benchmark} scene.
 }
  \label{fig:pieces}
\end{figure*}

\begin{figure*}[!t]
    \centering
    \subfloat[\label{fig:cases:a_exp01}]{
    \includegraphics[align=c,width=0.5\linewidth]{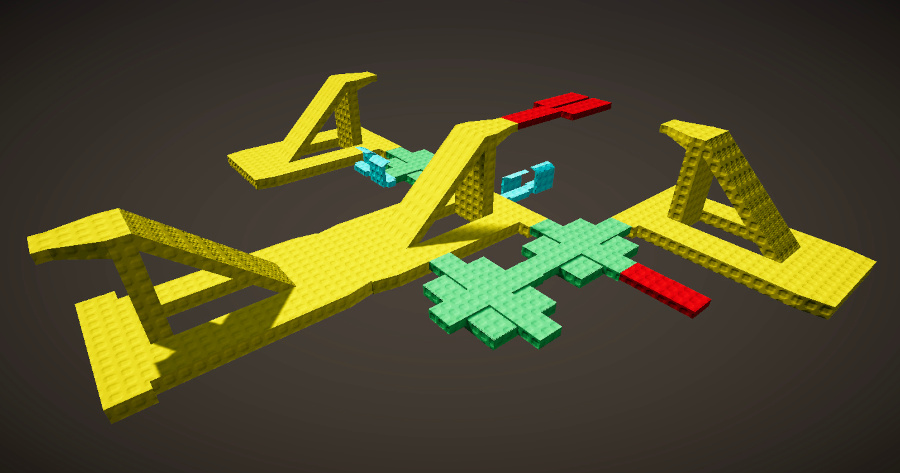}}
    \subfloat[\label{fig:cases:b_exp02}]{
    \includegraphics[align=c,clip,width=0.5\linewidth]{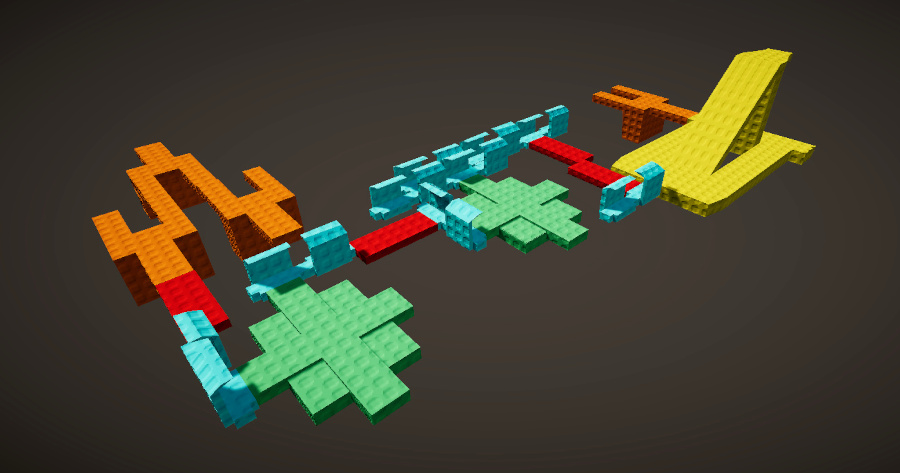}}

    \subfloat[\label{fig:cases:c_exp03}]{
    \includegraphics[align=c,width=0.5\linewidth]{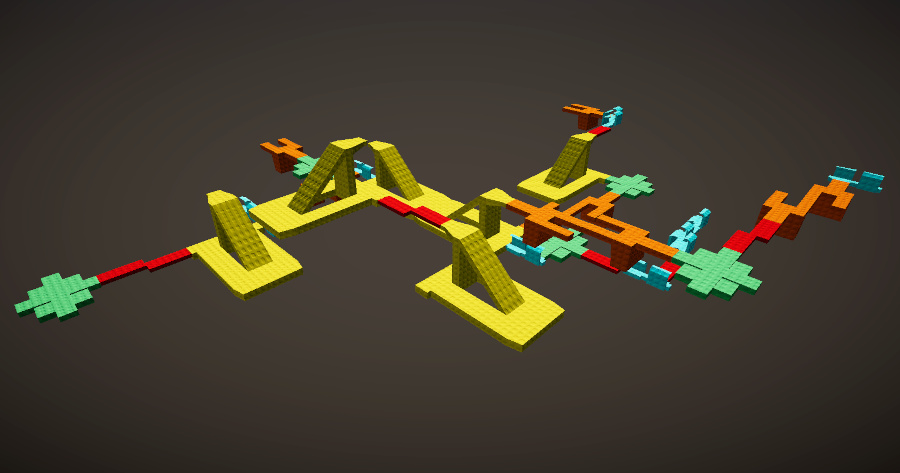}}
    \subfloat[\label{fig:cases:d_exp04}]{
    \includegraphics[align=c,clip,width=0.5\linewidth]{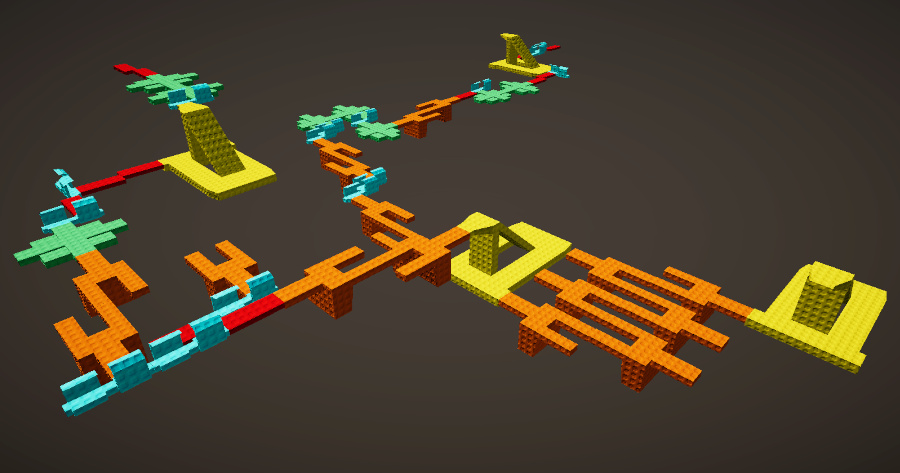}}

    \subfloat[\label{fig:cases:e_exp05}]{
    \includegraphics[align=c,width=0.5\linewidth]{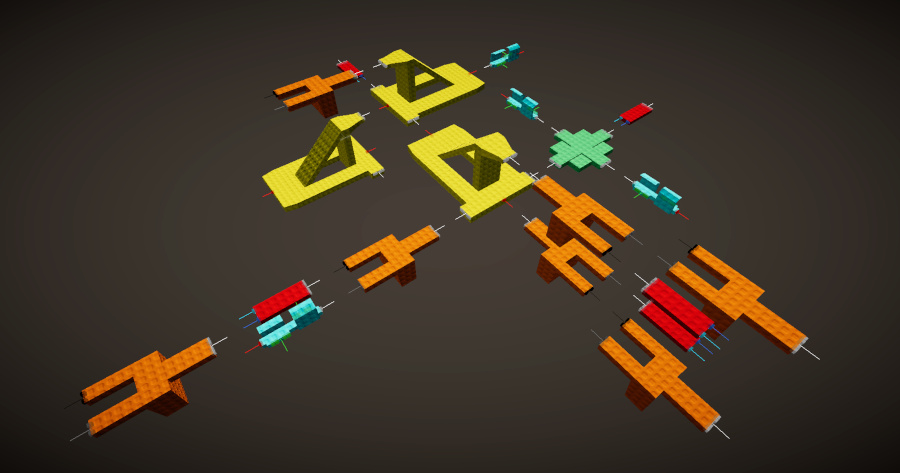}}
    \subfloat[\label{fig:cases:f_exp06}]{
    \includegraphics[align=c,clip,width=0.5\linewidth]{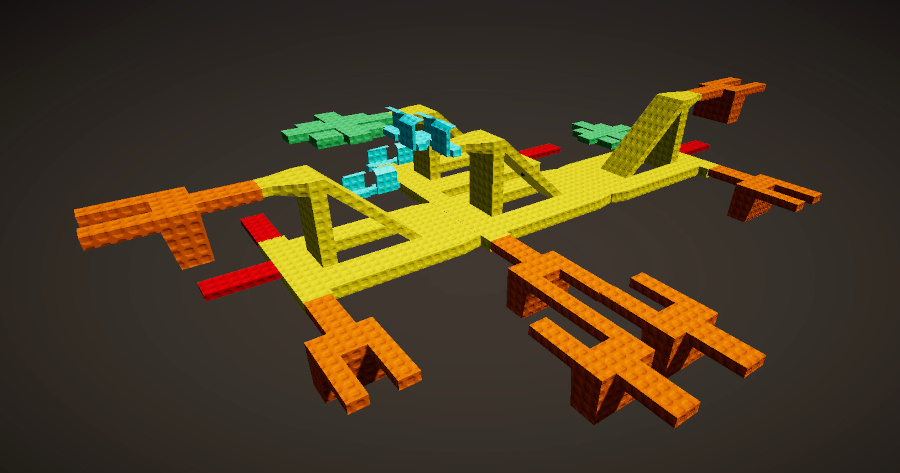}}

    \subfloat[\label{fig:cases:g_exp07}]{
    \includegraphics[align=c,clip,width=0.5\linewidth]{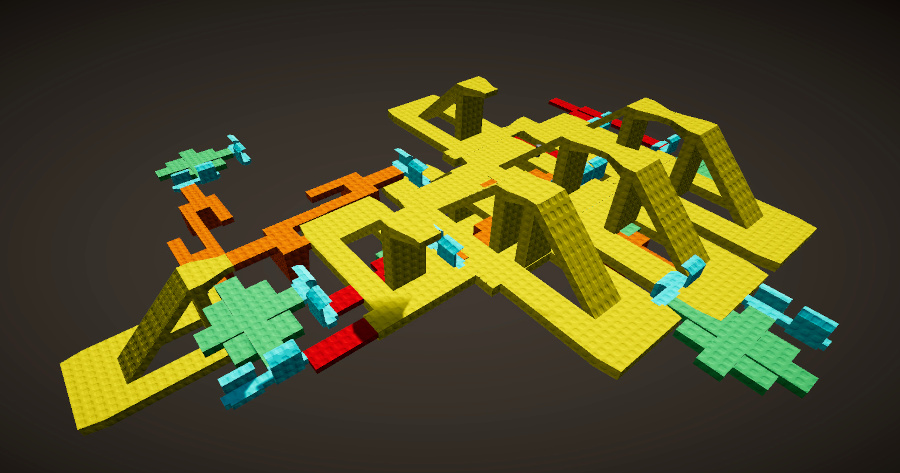}}
    \subfloat[\label{fig:cases:h_exp08}]{
    \includegraphics[align=c,clip,width=0.5\linewidth]{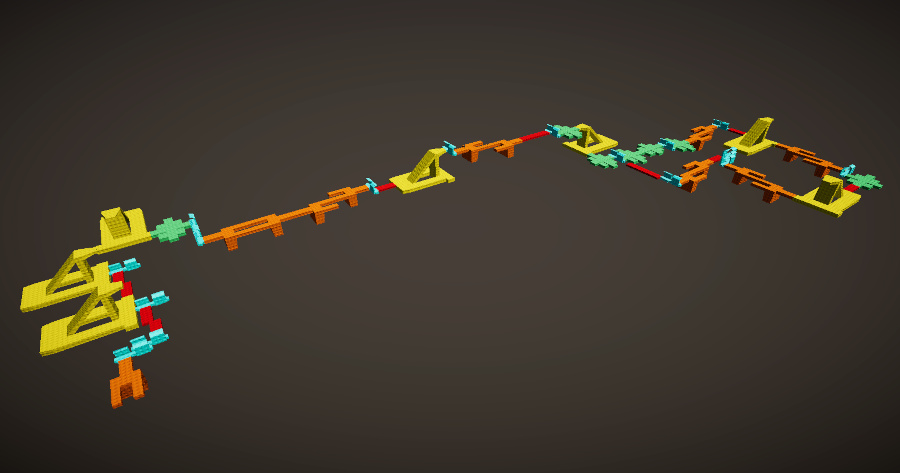}}

    \caption{Several illustrative maps generated with snappable meshes using the
    \textit{Benchmark} scene's map pieces presented in Fig.~\ref{fig:pieces}. The
    parameters used to generate each of these maps are given in Table~\ref{tab:cases}.}
    \label{fig:cases}
\end{figure*}

A number of experiments were performed in order to provide a better understanding of
the capabilities and limitations of the proposed technique. These experiments were
carried out using the \textit{Benchmark} scene, which contains the map pieces shown in
Fig.~\ref{fig:pieces}. Many different maps were generated during these experiments,
and several illustrative cases are shown in Fig.~\ref{fig:cases}, with their respective
parameters given in Table~\ref{tab:cases}.

\begin{table*}[!t]
  \centering
  \caption{Parameters Used to Generate the Maps Shown in Fig.~\ref{fig:cases}}
  \label{tab:cases}
  \begin{tabular}{lrlp{6.35cm}lrrrr}
  \toprule
  \multirow{2}{0.4cm}{Map} & \multirow{2}{0.6cm}{Piece count} & \multirow{2}{1.1cm}{Selection method} & \multirow{2}{*}{Selection method parameters} & \multirow{2}{1.5cm}{Matching rules} & \multicolumn{1}{l}{\multirow{2}{0.9cm}{Check overlaps}} & \multicolumn{1}{l}{\multirow{2}{0.65cm}{Max. fails}} & \multicolumn{1}{l}{\multirow{2}{1.2cm}{Piece distance}} & \multicolumn{1}{l}{\multirow{2}{*}{Seed}} \\
  & & & & & & & \\
  \midrule
  (a) & 13 & \textit{arena}    & \texttt{\scriptsize maxPieces=13} & Pins + Colors & \cmark & 10 & \num{0.0001} & $-267402550$\\
  (b) & 20 & \textit{corridor} & \texttt{\scriptsize maxPieces=20} & Pins + Colors & \cmark & 10 & \num{0.0001} & $-2095385667$\\
  (c) & 41 & \textit{star}     & \texttt{\scriptsize armLength=8, armLengthVar=2} & Pins + Colors & \cmark & 10 & \num{0.0001} & $277759099$ \\
  (d) & 56 & \textit{branch} & \texttt{\scriptsize branchCount=4, branchLength=12, branchLengthVar=4} & Pins + Colors & \cmark & 10 & \num{0.0001} & $1388449552$ \\
  (e) & 20 & \textit{arena}    & \texttt{\scriptsize maxPieces=21} & Colors\textsuperscript{a} & \cmark & 10 & \num{6.0000} & $811974397$ \\
  (f) & 21 & \textit{arena}    & \texttt{\scriptsize maxPieces=21} & Pins   & \cmark & 10 & \num{0.0001} & $-359152709$ \\
  (g) & 52 & \textit{star}     & \texttt{\scriptsize armLength=10, armLengthVar=4} & Pins + Colors & \xmark & 10 & \num{0.0001} & $1242840355$ \\
  (h) & 58 & \textit{corridor} & \texttt{\scriptsize maxPieces=120} & Pins & \cmark & 50 & \num{0.0001} & $-1444708658$ \\
  \bottomrule
  \multicolumn{9}{l}{\parbox[t]{0.97\linewidth}{\footnotesize{Notes: The `Piece count' column
  denotes the number of pieces effectively placed on the map. The \texttt{piecesList} parameter includes
  the five pieces shown in Fig.~\ref{fig:pieces}, with \texttt{useStarter} set to false. When applicable,
  \texttt{pinTolerance} is set to 0 and \texttt{colorMatrix} is as shown in Fig.~\ref{fig:params:connection}.
  The \texttt{starterConTol} selection method parameter is set to 0.}}}\\
  \multicolumn{8}{l}{{\footnotesize\textsuperscript{a}Connector colors are made visible in this example. Note that white connectors are configured as wildcards (see Fig.~\ref{fig:params:connection}).}}\\
   \end{tabular}
\end{table*}

Figs.~\ref{fig:cases:a_exp01}--\ref{fig:cases:d_exp04} display experimental maps created
with the different selection methods. A typical \textit{arena} method-generated layout
(\texttt{maxPieces=13}) is shown in Fig.~\ref{fig:cases:a_exp01}, where the map's tendency
to grow in all directions is clear. In turn, Fig.~\ref{fig:cases:b_exp02} shows a
map generated with the \textit{corridor} method (\texttt{maxPieces=20}). The
corridor characteristics are not immediately obvious, since, in this particular
example, each piece is being placed in a way that changes the direction established by
the previous piece. A larger map, created with the \textit{star} selection method, is
displayed in Fig.~\ref{fig:cases:c_exp03}, where the starting piece is a ``ramp''
(Fig.~\ref{fig:pieces:ramp}), which has five connectors. Therefore, and as expected,
the \textit{star} has five arms stretching out from the initial, center piece. Note
that one of these arms---the one expanding to the right in the figure---starts unfurling
at a higher altitude than the remaining arms. Finally, the map displayed in
Fig.~\ref{fig:cases:d_exp04} depicts the intended behavior of the \textit{branch}
selection method, with new ``branches'' created at specific points in the previously
placed pieces.

In Fig.~\ref{fig:cases:e_exp05}, the \texttt{pieceDistance} parameter was set to 6,
creating a map with several ``islands''. The connectors are shown to help visualize the
connections. The matching rules in this experiment were set to ``colors'', i.e., pieces snap
together based only on connector color, ignoring their pin count. The color matrix was set to
its default values (Fig.~\ref{fig:params:connection}), where connectors snap with connectors
of the same color, and white is set as a wildcard color, i.e., white connectors are able to
snap with connectors of any color. Changing the \texttt{pieceDistance} parameter in
this way generates mostly untraversable maps\footnote{Assuming characters without the
ability to jump or fly between areas.}, but allows for visual inspection or debugging,
for example to verify if connections are occurring according to the specified matching rules.

Contrary to the previous experiments, a pins-only matching rule was used to generate the
map displayed in Fig.~\ref{fig:cases:f_exp06}. As can be observed, the map has the
prototypical \textit{arena} layout, but the type of connections---without the color
matching constraint---are considerably different from those shown in
Fig.~\ref{fig:cases:a_exp01}.

An interesting experiment is shown in Fig.~\ref{fig:cases:g_exp07}, where the
overlap checking is disabled.
Since pieces are snapped together without consideration for collisions between them, the
resulting map loses the flow and clean appearance observed in the remaining experiments.
Disabling this option will, in most use cases, likely create map-wide geometry and texture
misalignments. Nonetheless, this may be desirable in some situations.

The goal in the last of these experiments was to create a long corridor-like map.
The \textit{corridor} selection method, with the \texttt{maxPieces} parameter set to
120, was used for this purpose. To avoid premature termination of the algorithm, a
pins-only matching rule was defined---thus eliminating the color matching constraint---and
the \texttt{maxFails} parameter was set to 50 (a value of 10 was used in the previous
experiments). The resulting map, displayed in Fig.~\ref{fig:cases:h_exp08},
is indeed a long corridor; however, the total number of pieces composing it is less than
half of \texttt{maxPieces}. This highlights the fact that, depending on parameter and
geometry constraints, the generated map may end up being smaller than envisaged by the designer.
Thus, it may be important to verify if the generated map attained some minimum threshold
concerning the number of placed pieces.

\subsubsection{Benchmarking the Example Maps}
\label{sec:casestudy:benchmarks:validation}

The illustrative maps generated with the \textit{Benchmark} scene, displayed in Fig.~\ref{fig:cases},
were benchmarked and analyzed from three perspectives: (1) the duration of the generation step; (2)
the time it takes to accurately execute the validation step; and, (3) the overall navigability of the
generated maps. The first two viewpoints clarify if the snappable meshes technique, as well as the
proposed validation step, allow for levels to be generated at runtime---especially given the quadratic
nature of the validation metrics. The third viewpoint demonstrates whether the snappable meshes
technique is overall able to generate valid, navigable maps. All experiments were performed with an
AMD Ryzen 7 5800X CPU, Ubuntu 20.04.3 LTS and Unity 2020.3.25f1 LTS.

\begin{figure}[!t]
  \centering
  \includegraphics[width=1\linewidth]{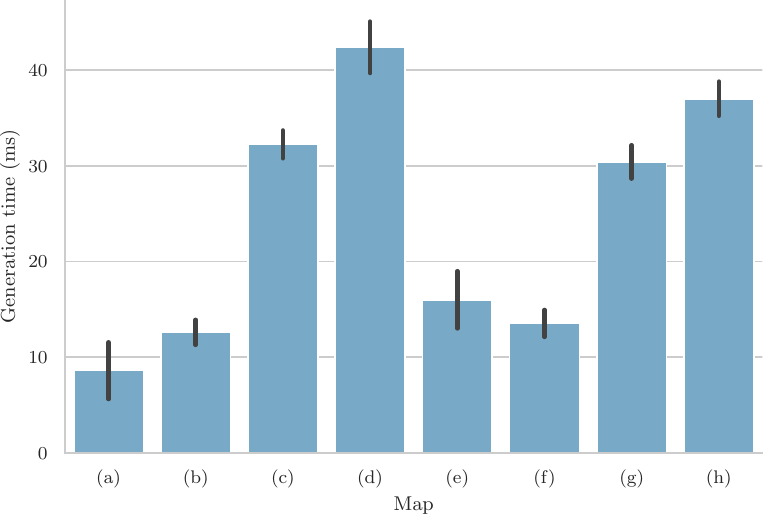}
  \caption{Mean map generation times in milliseconds, obtained by generating each map (a) to
  (h) 30 times. Error bars denote the sample standard deviation.}
  \label{fig:gentimes}
\end{figure}

The mean generation time was obtained by generating each map 30 times. It was observed to be as low as
\SI{8.6}{\ms} for map (a) (Fig.~\ref{fig:cases:a_exp01}), and up to \SI{42.4}{\ms} for map (d)
(Fig.~\ref{fig:cases:d_exp04}). Generation times for all the maps are displayed in
Fig.~\ref{fig:gentimes}. Even for maps of considerable dimensions, with over 50 pieces, the proposed
technique seems to be sufficiently fast to be used in a runtime level generation scenario.

\begin{figure*}[!t]
    \centering

    \subfloat[\label{fig:valid:50}]{
        \includegraphics[width=0.325\linewidth,align=c]{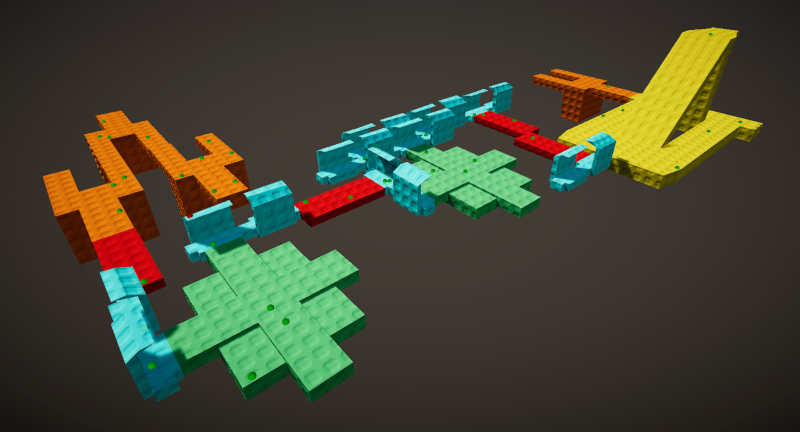}}
    \subfloat[\label{fig:valid:500}]{
        \includegraphics[width=0.325\linewidth,align=c]{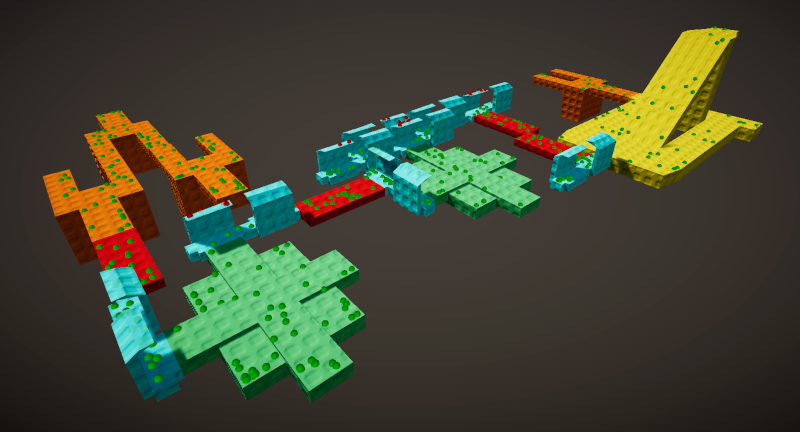}}
    \subfloat[\label{fig:valid:5000}]{
        \includegraphics[width=0.325\linewidth,align=c]{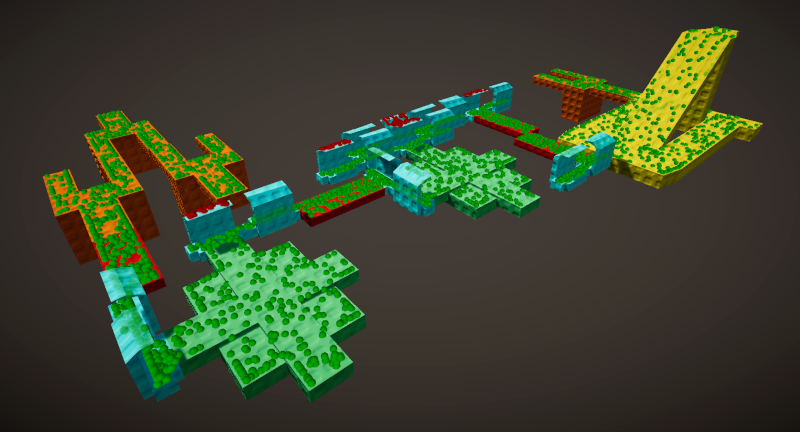}}

    \caption{Another perspective of the map in Fig.~\ref{fig:cases:b_exp02}, with the navigation
    points shown. Green navigation points belong to the largest fully-traversable region of
    the map, while red navigation points belong to other regions. (a) Using 50 navigation
    points; (b) using 500 navigation points; (c) using 5000 navigation points.}

    \label{fig:valid}
\end{figure*}

\begin{figure*}[!t]
    \centering
    {\scriptsize
    $n$ -- Number of navigation points\\
    \includegraphics[align=c,width=0.03\linewidth]{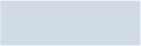} \parbox{0.03\linewidth}{$50$}
    \includegraphics[align=c,width=0.03\linewidth]{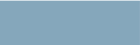} \parbox{0.03\linewidth}{$500$}
    \includegraphics[align=c,width=0.03\linewidth]{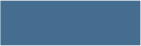} \parbox{0.03\linewidth}{$5000$}
    }
    \\
    \subfloat[\label{fig:results:c}]{
    \includegraphics[width=0.48\linewidth]{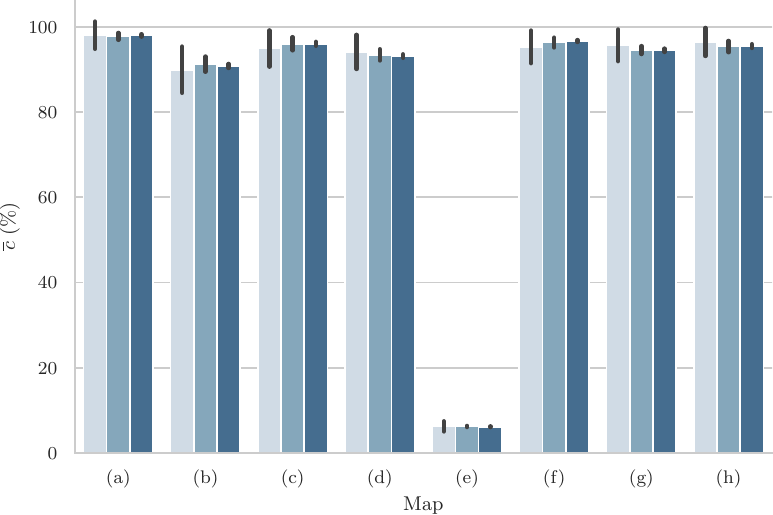}}
    \mbox{ }
    \subfloat[\label{fig:results:ar}]{
    \includegraphics[width=0.48\linewidth]{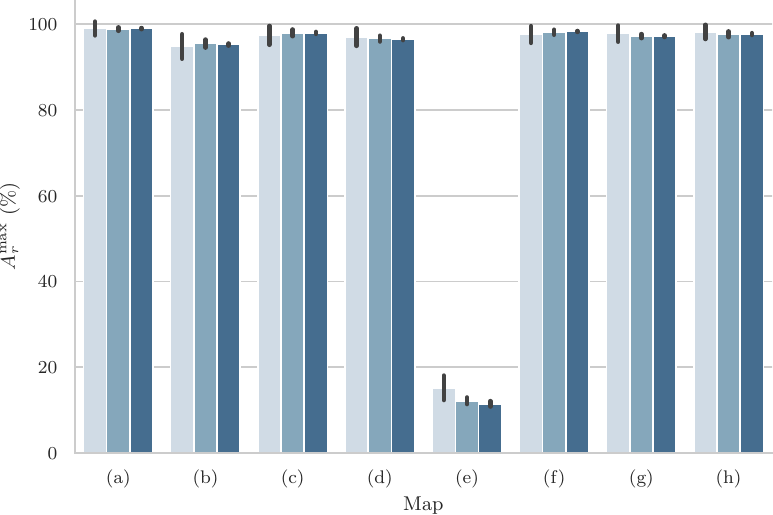}}

    \subfloat[\label{fig:results:nclu}]{
    \includegraphics[width=0.48\linewidth]{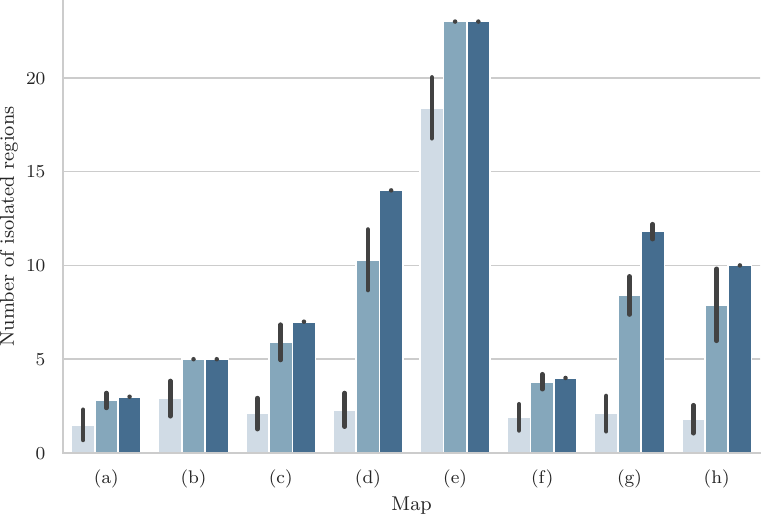}}
    \mbox{ }
    \subfloat[
    \label{fig:results:tv}]{
    \includegraphics[width=0.48\linewidth]{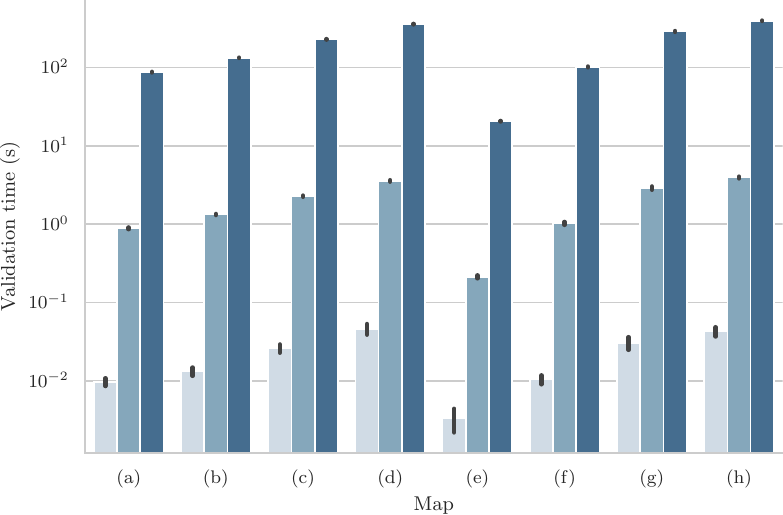}}

    \caption{Validation metrics obtained with 50, 500 and 5000 navigation points, for
    the maps shown in Fig.~\ref{fig:cases}, with parameters given in Table~\ref{tab:cases}.
    Bars display mean values obtained by validating each map 10 times, while error bars
    denote the sample standard deviation. Experiments were performed on the same
    hardware and software setup used for obtaining the generation times. The following metrics
    are presented:
    (a) average percentage of valid connections between navigation points, $\overline{c}$;
    (b) relative area of the largest fully-navigable map region, $A_r^\text{max}$;
    (c) number of isolated regions, i.e., of regions which are not connected to any another;
    and, (d) duration, in seconds, of the validation step---note the logarithmic scale.
    }
    \label{fig:casesvalid}
\end{figure*}

The validation step was performed with the number of navigation points, $n$, set to 50, 500 and
5000. An example of how these points are placed on a map is shown Fig.~\ref{fig:valid}, for the case
of map (b). The following data was collected over 10 runs for each combination of map and $n$:
$\overline{c}$ (average percentage of valid connections between navigation points),
$A_r^\text{max}$ (relative area of the largest fully-navigable region), number of regions which
are not connected to any other, and the duration of the validation step. Results are presented
in Fig.~\ref{fig:casesvalid}, while the raw data and respective analysis are available online
\cite{fachada2022smdata}.

The quadratic nature of the validation step, explained in Subsection~\ref{sec:methods:valid},
is made obvious from the results shown in Fig.~\ref{fig:results:tv}: a $10\times$ increase in
the number of navigation points leads to an approximately $100\times$ longer validation step.
While the duration of the validation step for 50 navigation points---in the order of a few
milliseconds---is certainly acceptable for runtime map generation (and likely feasible with
500 points, with validation times of a few seconds at most), that is clearly not the case for
5000 points, where validation can take up to several minutes. The question here is whether
there is a significant difference in validation accuracy when using considerably more navigation
points. As Fig.~\ref{fig:results:c} and Fig.~\ref{fig:results:ar} show, validation results
are quite similar when deploying 50, 500 or 5000 points, with no clear tendency for an
increase or decrease in percentage for both the $\overline{c}$ and $A_r^\text{max}$ metrics.
Thus, it seems possible to conclude that a relatively low amount of navigation points
produces sufficiently accurate validation metrics, while being fast enough for runtime map
generation. In turn, the number of detected low-navigability areas increases with $n$
(Fig.~\ref{fig:results:nclu}), which is to be expected, since a higher point coverage boosts
the chances of finding these small, isolated regions. However, this is not relevant for runtime
map generation, since the main concern there is in finding the largest navigable region, and
this is done successfully even with very few points, as highlighted in Fig.~\ref{fig:results:ar}.

With respect to the overall navigability of the generated maps, results hint that the
proposed method is robust, yielding highly-traversable maps for a variety of
parameterizations. We have performed a large number of additional experiments, with
various parameters and seeds, and have observed that $\overline{c}$ is rarely below
\SI{90}{\percent}, while $A_r^\text{max}$ is typically above \SI{95}{\percent}. This is
of course assuming sensible parameterizations. For example, map (e), shown in
Fig.~\ref{fig:cases:e_exp05}, has very poor navigability due to the large
\texttt{pieceDistance}. An interesting aspect in these validation experiments, which can
be observed in the red navigation points in Fig.~\ref{fig:valid}, is that the
low-navigability areas in the generated maps are essentially limited to the rooftops of
the ``hallway'' piece (Fig.~\ref{fig:pieces:hallway}). Therefore, it is clear that piece
design has considerable influence on validation metrics, and that the results presented
here, although showing promise for general use cases, should be mainly considered in the
context of the utilized pieces.

\subsection{Exploring the Generative Capabilities of Snappable Meshes}
\label{sec:casestudy:altpieces}

In this second case study, we evaluate four maps generated with the \textit{Artistic} scene,
also bundled with the Unity prototype implementation. This scene is configured with a set of
pieces notably distinct from those used in the previously discussed \textit{Benchmark} scene,
featuring cleaner and seamless connection interfaces. The scene includes 6 ``hub''-like pieces
and 6 ``corridor''-like pieces, some of which have stairs or ramps, allowing for multistory maps.
All 12 pieces are specified in \texttt{piecesList}. However, the \texttt{starterList} contains
only the ``hub'' pieces (thus, \texttt{useStarter} is set to \texttt{true}). The ``hub'' pieces
have 4 or 5 connectors with 2 pins each, while the ``corridor'' pieces have 2 or 3 connectors with
2 or 3 pins each. All connectors are white, thus connections are made only by pin count, with
\texttt{pinTolerance} set to zero. Therefore, ``hubs'' can snap with other ``hubs'' without
restriction (other than having free connectors), while ``corridors'' enforce some constrains on
what connections are possible. In particular, some ``corridor'' connectors can only snap to
compatible connectors on other ``corridor'' pieces, while some other ``corridor'' connectors can
also connect to ``hubs''. The motivation for this design was in creating maps with larger areas
connected with ``corridor''-like pieces.

\begin{figure*}[!t]
    \centering

    \subfloat[\label{fig:altmaps:arena}]{
        \includegraphics[width=0.48\linewidth,align=c]{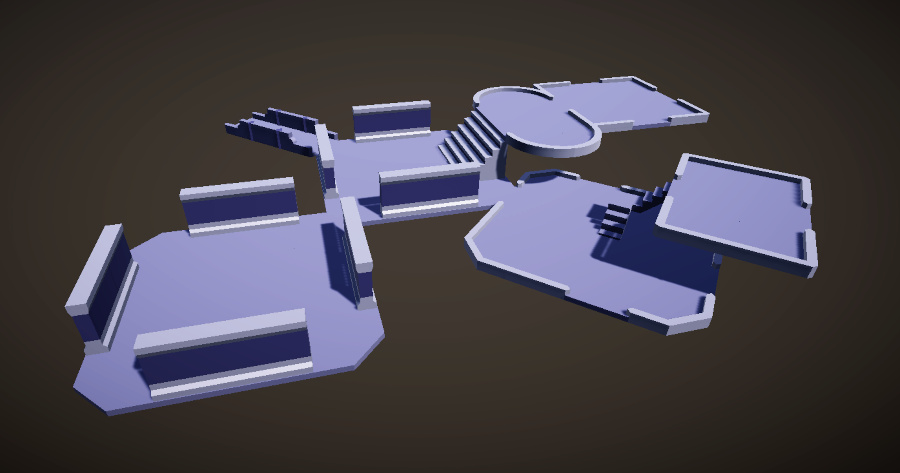}}
    \subfloat[\label{fig:altmaps:corridor}]{
        \includegraphics[width=0.48\linewidth,align=c]{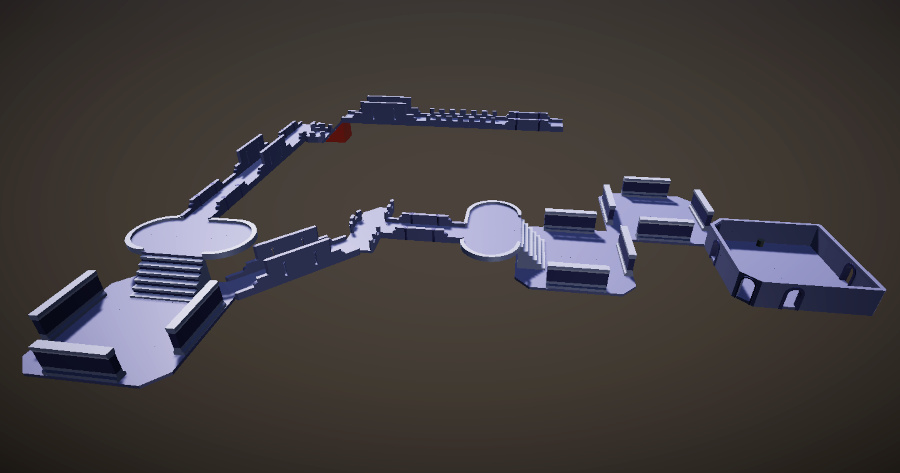}}
    \\
    \subfloat[\label{fig:altmaps:star}]{
        \includegraphics[width=0.48\linewidth,align=c]{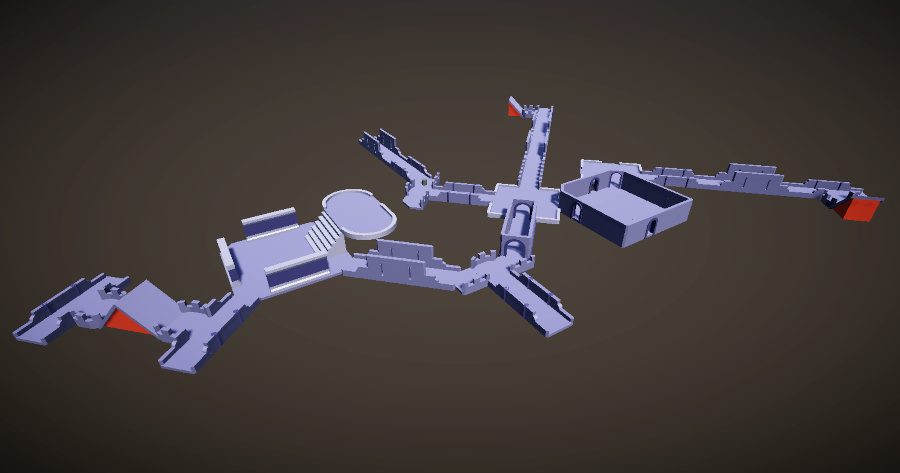}}
    \subfloat[\label{fig:altmaps:branch}]{
        \includegraphics[width=0.48\linewidth,align=c]{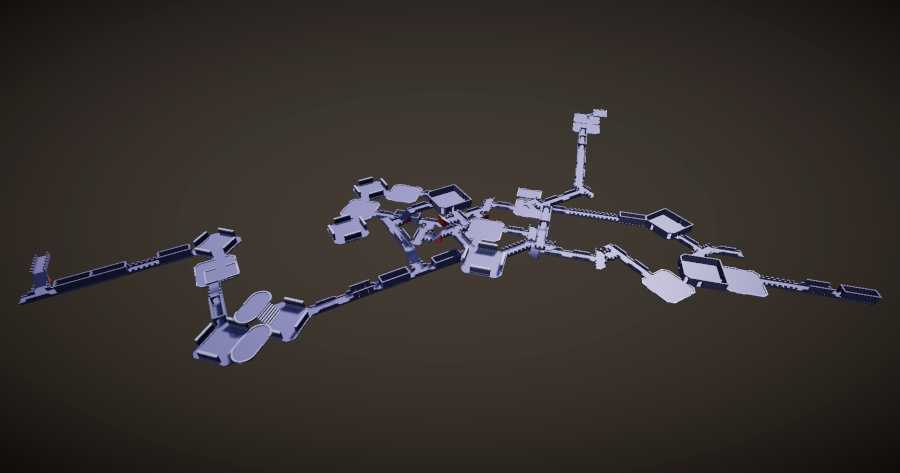}}

    \caption{Example maps created under the \textit{Artistic} scene in the Unity prototype
    implementation.
    (a) Using the \textit{arena} selection method;
    (b) using the \textit{corridor} selection method;
    (c) using the \textit{star} selection method;
    (d) using the \textit{branch} selection method.}

    \label{fig:altmaps}
\end{figure*}

The generated maps, shown in Fig.~\ref{fig:altmaps}, were created with the \textit{arena},
\textit{corridor}, \textit{star}, and \textit{branch} selection methods, respectively. In all
instances, \texttt{pieceDistance} and \texttt{checkOverlaps}---relevant parameters for this
discussion---were set to \num{0.0001} and \texttt{true}, respectively. The complete configuration
for these examples is predefined in one of the \textit{Artistic} scene's panels, making them easily
reproducible. The four maps are 100\% navigable---essentially due to piece design---and
generation/validation times are very similar to what was observed in the case of the
\textit{Benchmark} scene. Therefore, we will focus this analysis on the maps' aesthetic properties
and their gameplay characteristics, highlighting the diverse generative possibilities put forth by
the snappable meshes algorithm.

The first map generated with the \textit{Artistic} scene, shown in Fig.~\ref{fig:altmaps:arena},
was created with the \textit{arena} selection method, with \texttt{maxPieces} set to 18, although
only 5 pieces---4 ``hubs'' and 1 ``corridor''---are actually placed on the map before the algorithm
terminates. The map is essentially a single concentrated area, typical of the \textit{arena}
selection method, and offers a number of open areas, cover positions and vantage points. Due to
piece design, the map presents a more organic and playable style than the maps shown in
Fig.~\ref{fig:cases}, while stairs in two ``hub'' pieces promote the mentioned vantage points.

The second example, presented in Fig.~\ref{fig:altmaps:corridor}, consists of a map created with the
\textit{corridor} selection method. This is a linear map, with clear extremities. The stairs in two
of the ``hub'' pieces allow the map to have varying altitudes along its linear path. The
\texttt{maxPieces} parameter was set to 12, and this is the number of building blocks effectively
placed on the map. The starting piece is the ``hub' to the right edge of the figure, somewhat counter
to how the \texttt{corridor} method works, since it chooses the block with least connectors as the
starting piece. Note, however, that in the \textit{Artistic} scene, the \texttt{starterList} only
contains ``hub'' pieces, emphasizing the flexibility of the snappable meshes technique.

Fig.~\ref{fig:altmaps:star} demonstrates the prototypical \textit{star}-generated layout, showing a
basic center ``hub'' piece with four arms opening up in all directions. Naturally, the \textit{star}
selection method was used to create this map, with \texttt{armLength} and \texttt{armLengthVar} set
to 5 and 2, respectively. The actual length of the four arms is, from the perspective of the center
piece, 3 (left), 4 (top), 6 (right), and 5 (bottom), which, together with the center piece, yield a
map with a total of 19 building blocks. Two of the arms are essentially ``corridors'', while the other
two also contain ``hub'' pieces.

The \textit{branch} selection method was used to create the fourth and last example, shown in
Fig.~\ref{fig:altmaps:branch}. The map displays several larger areas composed of ``hubs'', connected
by ``corridors'', as intended in the piece design. The branching effect provides a non-linear level
experience with multiple areas---appropriate for exploration, for example. The map has a total of 70
pieces, and the selection method parameters, \texttt{branchCount}, \texttt{branchLength}, and
\texttt{branchLengthVar}, were set to 5, 12, and 2, respectively.

These examples offer a broader picture of the proposed technique's generative capabilities. An
important aspect that stands out in these four maps is their free-flow nature, which would be
difficult to achieve with grid-based map PCG methods. Additionally, and even though the
algorithm does not explicitly consider multistory levels, it is still possible to add a sense
of verticality with appropriate piece design, further underlining the technique's versatility, as
well as its reliance on properly constructed building blocks.

\subsection{\textit{Trinity}---a Third-Person Multiplayer Shooter}
\label{sec:casestudy:trinity}

\begin{figure}[!t]
  \centering
  \includegraphics[width=0.9\linewidth]{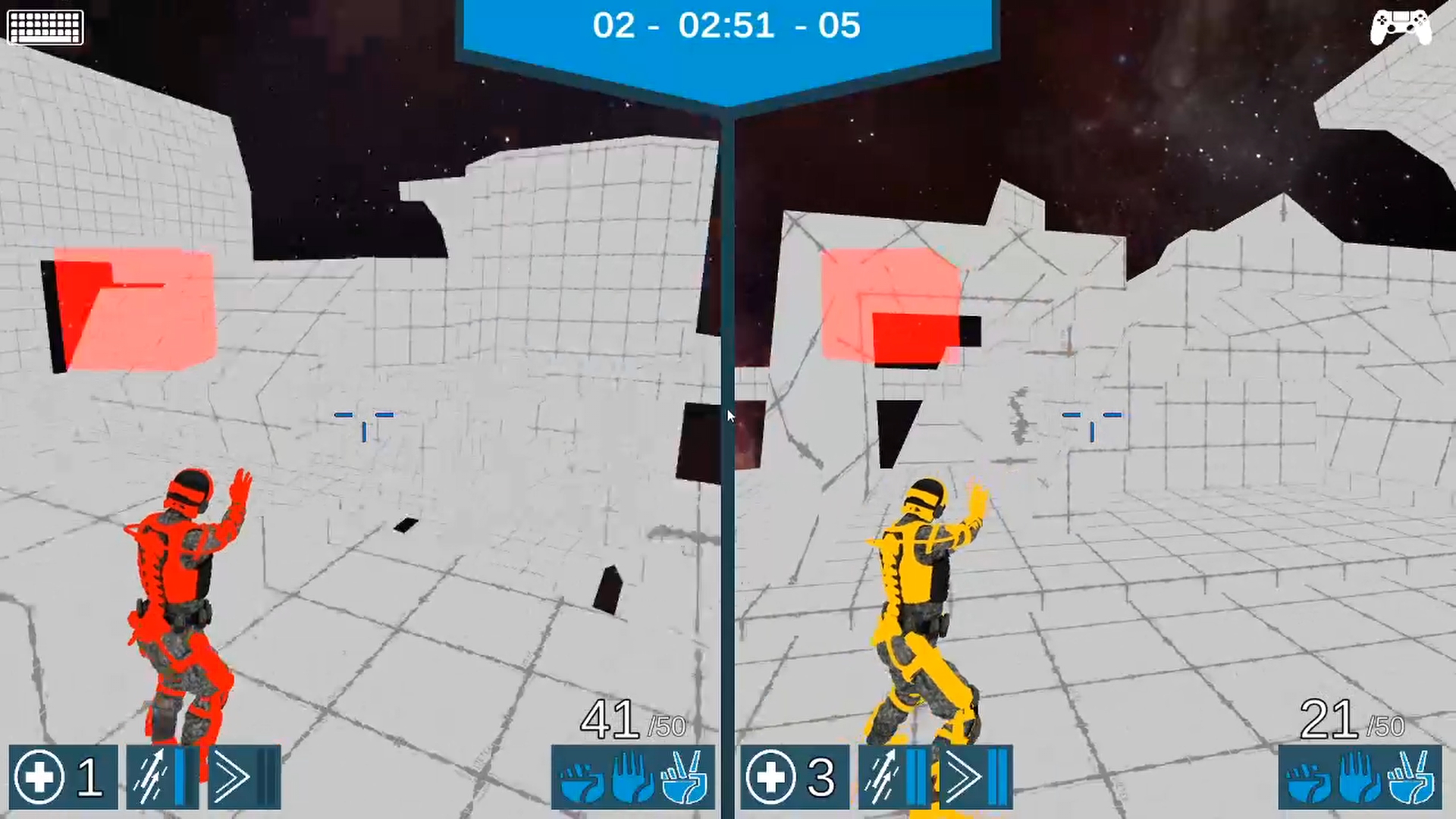}
  \caption{Screenshot of \textit{Trinity} during a match.}
  \label{fig:ingame}
\end{figure}

\textit{Trinity} is a competitive, split-screen multiplayer game
(Fig.~\ref{fig:ingame}), developed as a semester project at Lusófona University's Bachelor
in Videogames \cite{fachada2020topdown}. It is a third-person shooter in which players
navigate the environment trying to eliminate their opponents using weapons that shoot
different types of ammunition with various effects and counter-effects.

The snappable meshes algorithm was initially developed to create maps for this game,
generating a new procedural playfield at the start of each match. The method allowed
designers to quickly experiment with many different map types during the development
stage, and, together with testers, to swiftly home in on
a set of map pieces and algorithm parameters deemed suitable for the goals set forth
for \textit{Trinity}. In the final game, a single set of map pieces and parameters were
selected for generating maps. The \textit{arena} selection method was chosen since it
created satisfactory layouts---typically large areas with both open spaces and plenty of
options for platforming and cover. The \textit{corridor} method was also considered.
It was better suited for ``capture the flag''-style matches due to its tendency to
create layouts that flowed along one direction. However, this game mode was not
implemented, thus \textit{corridor} generation was left out of the final version of the
game.

Although limited to one parameter set in the final game, the technique enhanced the game's
replayability. More specifically, the procedurally generated levels changed the
conventional play style of shooter games, since in \textit{Trinity}, players have to quickly
adapt to the playfield, instead of memorizing and using prior knowledge of the levels.

\section{Discussion}
\label{sec:discussion}

In this section we start by contextualizing snappable meshes as a designer-centric
approach, framing our reasoning with recent literature
(Subsection~\ref{sec:discussion:framing}). A discussion of the technique's use
cases is undertaken in Subsection~\ref{sec:discussion:uses}, where its potential
as a prototyping system is highlighted. Subsection~\ref{sec:discussion:lims}
points out a number of limitations in this study, namely at the level of the
proposed method and the provided prototype implementation, as well as their
evaluation. Finally, a number of alternative uses and possible improvements are
explored in Subsection~\ref{sec:discussion:future}.

\subsection{Snappable Meshes as a Designer-centric Approach}
\label{sec:discussion:framing}

Togelius et al. \cite{togelius2016intro} put forward five desirable properties of a PCG
solution, namely \textit{speed}, \textit{reliability}, \textit{controllability},
\textit{expressivity/diversity}, and \textit{creativity/believability}. As shown in the
previous section, the snappable meshes algorithm is relatively \textit{fast}, with
generation times in the order of milliseconds, and allows the general look and feel of
the generated maps to be \textit{controlled}, although not their exact layout.
The method is also able to provide \textit{reliable} maps---from a
traversability perspective, at least---if the generated maps are validated with the
criteria defined in Subsection~\ref{sec:methods:valid}. Since the technique requires map
pieces to be created and provided by the designer, it has a high \textit{expressivity} and
\textit{diversity} potential, being able to generate distinct maps given different building
blocks. However, for this same reason, the snappable meshes technique cannot, by itself,
guarantee \textit{creative} and \textit{believable} maps. If given poorly designed pieces,
the method will likely produce maps that are neither.

According to Zhu et al. \cite{zhu2018explainable}, a common issue with AI techniques in
general and PCG methods in particular is that of increasing algorithmic complexity,
which hinders the designer's understanding and trust about what the algorithm is doing.
Consequently, designers are likely to avoid using such techniques to their full potential
or not use them at all. A possible solution or mitigation for this problem is to develop,
from the ground up, designer-oriented and fully explainable PCG methods and techniques
\cite{zhu2018explainable}. The proposed snappable meshes technique was outlined with these
considerations in mind. It has a straightforward and designer-centric workflow, and is
fully explainable, as shown by the generated logs in the Unity prototype implementation.
As suggested by Zhu et al. \cite{zhu2018explainable}, the generative process is narrated
in the form of a sequential textual description, in which the algorithm's decisions are
explained.

Snappable meshes imposes few restrictions. Issues can generally be solved via editing
map pieces, changing the algorithm parameters, or, if necessary, by creating new selection
methods. Further, it provides immediate results, avoiding lengthy or computationally costly
procedures common in search-based \cite{togelius2011search} and interactive evolution
approaches, which may result in user fatigue \cite{liapis2016mixed}, as well as in
difficulties in fully understanding how to control the generation process
\cite{zhu2018explainable,baldwin2017towards,almeida2013systematic}. In edge cases, new
selection methods could implement or incorporate search- or machine learning-based
strategies in a compositional fashion \cite{togelius2012compositional}---though this would
undermine the simplicity and speed offered out of the box by the proposed method.

In addition, the snappable meshes technique aims to respect the designer according to the
three pillars set forth by Lai et al. \cite{lai2020towards}, namely (1) \textit{respect
designer control}, (2) \textit{respect the creative process}, and, (3) \textit{respect
existing work processes}. In regard to pillar 1, the proposed technique respects designer
control since, as already discussed, it provides ``enough control to bring out the designer's
vision'' \cite{lai2020towards}. Pillar 2, respecting the creative process, ``concerns itself
with having a feedback loop that is short enough that the creative process is not disturbed''
\cite{lai2020towards}. This is guaranteed by the technique's short generation and validation
time. Finally, the snappable meshes approach respects existing work processes, the third pillar,
since the algorithm can be integrated in existing workflows (e.g., game engines), and make use
of existing assets---although these need to be ``decorated'' with connectors. The validation
metrics, discussed in Section~\ref{sec:methods:valid}, offer yet another perspective of
respect for the designer, as they cater for designers mainly interested on why a map is ``deemed
unplayable by the AI agent'' \cite{zhu2018explainable}.

Considering the increasing rift between
academia and industry related to communication and used methodologies \cite{lai2020towards},
we believe the simplicity, explainability, extensibility and respect for the designer embedded
in the proposed technique grants it the potential to reduce this gap.

\subsection{Use Cases}
\label{sec:discussion:uses}

The most obvious use case for the snappable meshes technique is the one it was
initially developed for: generating maps for 3D multiplayer shooters, possibly in a
generate-and-test loop to guarantee adequately navigable levels. This is, however,
a reductionist view of the technique's potential, since there is nothing limiting
its use in other game genres or scenarios. On the contrary: since the fundamental
blocks of a map are created by the human designer, and given the possibility of
adding new selection methods, snappable meshes can be considered more of a meta-PCG
approach rather than a concrete algorithm or tool for specific use cases. However,
framing the technique on such general terms is also not helpful, as one might be
tempted to simply state that it can eventually create anything. Therefore, we
will highlight an important use case, not related to any particular game type or
genre: using snappable meshes as a prototyping method or visual map design approach
\cite{almeida2013systematic}, appropriate to designers with little to no programming
experience, boosting their design space \cite{noor2016evaluating}, allowing for fast
iteration and speeding up development.

Particularizing on the use of the proposed technique as a visual prototyping method,
connectors can be used as measuring aids for spacing in the game world. A designer can
easily define a passage as $n$ pins wide or tall, keeping consistency in the design of
the layout of the individual pieces being made separately. Not only can a snappable
meshes implementation be used as a level creation tool, but the generated maps can also
be used to kickstart discussions between collaborating developers and to create basic
rules for the construction of pieces, even if the algorithm ends up not being used in
the final designs. Further, by using the color matching rules, the pieces developed by
one designer can be grouped in the final outputs, allowing for focused design and
prototyping of pieces belonging to specific areas or sections that can be seamlessly
combined together while keeping a mixed-authorship approach (both human-human and
human-computer) throughout the whole process.

\subsection{Limitations}
\label{sec:discussion:lims}

While the proposed approach frees the designer of grid and space restrictions
when creating map pieces, care must be taken in their design in order to
maintain cohesion and predictability of the generated output. In other words,
and highlighting what was stated in Subsection~\ref{sec:discussion:framing},
the algorithm will most likely produce poor maps given substandard building
blocks.
Piece design becomes even more important considering that, to promote
simplicity and explainability, the algorithm works in a greedy fashion---i.e.,
the first random piece that ``fits'' is selected---and does not perform
backtracking. Since the search space is not thoroughly explored in a single
algorithm run, this may lead to weak configurations and/or early dead ends, an
issue potentially aggravated by poor piece design. Nonetheless, this
limitation is offset by the technique's short generation times, which allow it
to work in a generate-and-test loop with validation metrics such as the ones
discussed in Section~\ref{sec:methods:valid}. Thus, the algorithm can be
executed multiple times, exploring the search space until a map with the desired
qualities is found.

Again, due to the focus on simplicity and explainability, there is no planning
or any kind of spatial analysis when placing new pieces on the map, which would
be necessary for creating loops and faultlessly avoiding the dead ends mentioned
in the previous paragraph. Disabling overlap detection solves both problems at the
cost of elegant map designs, but this is far from an ideal solution.

Another issue with the snappable meshes technique concerns the navigability
validation presented in Subsection~\ref{sec:methods:valid}, where we proposed
deploying a predetermined amount of randomly distributed navigation points on
the map, and then verify their connectivity. In the prototype Unity
implementation these points are randomly placed in the runtime-generated
navmesh. As shown in Fig.~\ref{fig:valid}, some of these points may be
placed on rooftops or other areas not intended for navigation; thus, if this
approach is followed as-is for indoor maps with hollow map pieces, it may lead
to invalid paths outside the intended play area. Consequently, individual map
pieces may require additional metadata specifying valid movement zones.

Another limitation in the provided Unity prototype, also related with navigation,
is the fact that it currently does not support jumps. While this is essentially an
implementation detail---nothing stops an improved implementation of providing this
functionality---it can hinder a more thorough experimentation by interested readers.

The generality of snappable meshes as a meta-PCG technique, as opposed to more
objective PCG tools, creates some difficulties pertaining to runtime validation and
comprehensive method evaluation. With respect to runtime validation, it is not feasible
to exhaustively test or validate generated maps without knowing the specific context in
which they will be deployed. As such, no validation metrics are presented in this
paper other than navigability. This is the same reason why a comprehensive method
evaluation is not performed in this paper. Such evaluations offer quality assurances
on the generated content, and typically involve a top-down statistical analysis of the
technique's generation space \cite{noor2016evaluating}. An analysis of this kind is
difficult to perform on an open-ended method such as snappable meshes, which, as already
stated, is essentially a meta-generator, heavily dependent on the designer-provided
blocks and lacking a predefined goal on the type of maps to generate. Thus, the
difficulties in obtaining general statistical assurances on the generated content are
in effect a limitation of the snappable meshes technique.

\subsection{Alternative Uses and Possible Improvements}
\label{sec:discussion:future}

Given the open-endedness of the proposed technique, alternative or unanticipated
uses are possible within the presented framework simply with smart and/or
creative map piece design, as well as by implementing new generation methods.

An interesting possibility would be to use connector constraints (e.g., connector
color constraints) to combine not only full map pieces, but also props, obstacles, and
even simple cosmetic changes to those same pieces---possibly even player characters
or NPCs. The possibility of creating more ``final'' maps also opens the door for
validation metrics beyond navigability, such as cover ratios, target visibility,
or detection of dangerous hotspots. Level difficulty could be assessed by
determining path costs taking enemies and obstacles in consideration, as done by
Togelius et al. \cite{togelius2012compositional}, for example.

An innovative avenue of research would be to combine the input rearrangement and
look-ahead capabilities of WFC with the free-form, gridless approach taken by
snappable meshes. This would allow the creation of elegant and elaborate 3D levels
given a single input example, such as levels carefully crafted by human designers
or levels from existing games. Further, WFC's more general constraint solving ability
could potentially lead to more plausible environments with stricter constraints from
a gameplay perspective.

Looking outside of the presented framework, the algorithm may be extended by
performing multiple generation passes. This could potentially produce maps with
several floors, adding explicit verticality to the designs, and foster more complex
layouts by allowing the designer to specify different parameters for each pass.
Additional passes could also be used to connect large separated clusters, for example.

\section{Conclusions}
\label{sec:conclusion}

In this paper we presented the snappable meshes PCG technique for creating
gridless 3D maps based on designer-modeled meshes with visual connection
constraints. The approach was thoroughly described from an algorithmic perspective,
and a Unity prototype implementation was introduced as a practical way of
experimenting, testing, and studying the method. The case studies discussed in
Section~\ref{sec:casestudy} showed that the technique can be used for generating
levels in a concrete game scenario, and is able to generate a variety of map layouts,
even with a limited set of building blocks. Further, both map generation and validation
were shown to be fast procedures, opening the door for using snappable meshes in
runtime generate-and-test loops. We argued that the technique respects the designer,
offering a degree of control on the look and feel of the generated maps, while being
adaptable to existing workflows. We also highlighted the potential of snappable
meshes as a collaborative prototyping methodology, while discussing its limitations,
alternative use cases and possible algorithmic improvements. In sum, the proposed
method was shown to be a viable map creation solution, allowing  fast and/or
collaborative level design and prototyping.

\section*{Acknowledgment}

The authors would like to thank Phil Lopes for his valuable suggestions.
This work was supported by Fundação para a Ciência e a Tecnologia under
Grant No.: UIDB/04111/2020 (COPELABS).

\ifCLASSOPTIONcaptionsoff
  \newpage
\fi

\bibliographystyle{IEEEtran}

\end{document}